\documentclass[10pt]{article}
\usepackage[preprint]{tmlr}

\usepackage{hyperref}
\usepackage{url}
\usepackage{graphicx}
\usepackage{amsmath,amssymb}
\usepackage{booktabs}
\usepackage{array}
\newcolumntype{P}[1]{>{\raggedright\arraybackslash}p{#1}}
\usepackage{tikz}
\usetikzlibrary{arrows.meta,calc}
\definecolor{repblue}{HTML}{0072B2}
\definecolor{loopverm}{HTML}{D55E00}
\definecolor{credgray}{HTML}{3A3A3A}
\definecolor{storefill}{HTML}{FFF3CC}

\title{When Does Backpropagating Through Policy Memory Matter? Separating Representation Credit from Closed-Loop Credit}

\author{
    \hspace*{-0.4em}\name Xingjian Li$^{1}$ \email 123090292@link.cuhk.edu.cn
    \AND
    \name Jianhua Z. Huang$^{1,2}$ \email jhuang@cuhk.edu.cn
    \AND
    \name Junli Duan$^{3,\dagger}$ \email duan.junli@coscoshipping.com
    \AND
    \addr $^1$  School of Data Science, The Chinese University of Hong Kong, Shenzhen \\
    \addr $^2$ School of Artificial Intelligence, The Chinese University of Hong Kong, Shenzhen \\
    \addr $^3$ COSCO Shipping Technology Co., Ltd. \\
    \addr $^\dagger$ Corresponding author.
}

\newcommand{\sg}{\mathrm{sg}}

\def\month{10}
\def\year{2026}

\begin{document}

\maketitle

\begin{abstract}
When a policy with memory is trained through a differentiable simulator, credit can reach what it stores along two routes: \emph{representation credit} teaches the network that writes the memory what to summarize, while \emph{closed-loop credit} travels back through the physical states that earlier actions produced and that the memory later stored. Stopping gradients at stored history, as truncated backpropagation and Transformer-XL do, removes both routes at once. Keeping every forward computation identical and removing one route at a time, we trained Transformer policies for quadrotor tracking to determine when each route matters. Credit through memory improved learning only when velocity observations were noisy or hidden. When velocity was hidden, mildly decaying the gradient between consecutive physical states ($\gamma = 0.8$) eliminated the truncation cost of removing closed-loop credit while leaving representation credit essential, a separation that held in 14 of 16 new seeds. Two further systems bound this result: in a fixed-wing aircraft, representation credit could replace closed-loop credit on its own, leaving no cost to separate, and in a cart-pole, training with full state-edge credit was too unstable to compare the routes. In both the quadrotor and a fully observed vessel model, static gradient norms failed to predict which cut cost more, so memory truncation choices are best guided by short training runs.
\end{abstract}

\section{Introduction}

Policies that act from partial observations rely on memory, and an increasing number of them are trained by differentiating through a physics simulator. Analytic policy gradients backpropagate the task loss through the simulated dynamics to the policy parameters \citep{wiedemann2023apg,xu2022shac}, and recurrent policies trained this way have been used for agile quadrotor flight \citep{zhang2025agile} and autonomous driving \citep{nachkov2025controllers}. Training such a policy raises a design choice that sequence modeling has long faced: whether to stop gradients at stored history. Truncated backpropagation through time cuts the recurrent graph after a fixed number of steps \citep{williams1990efficient}, and Transformer-XL reuses cached hidden states without backpropagating into them \citep{dai2019transformerxl}, a placement that memory-based reinforcement learning agents adopted as well \citep{parisotto2020gtrxl}. In a differentiable simulator, however, the cut removes more than internal network paths: a stored observation depends on the physical states that the policy's own earlier actions produced, so credit reaching memory can also flow back through those states.

Practitioners consequently face two truncation decisions that are usually made separately. The first is how far to backpropagate through the simulator, where shorter, adaptive, or decayed physical horizons have each been proposed to stabilize training \citep{xu2022shac,georgiev2024ahac,zhang2025agile}. The second is whether to backpropagate through policy memory, where practice usually follows sequence modeling on static datasets with no environment to differentiate through. This paper asks when backpropagating through policy memory matters for learning, and through which backward route that learning signal acts.

Answering this question takes more than comparing a policy with memory to a memoryless baseline, because removing memory changes what the policy can read in the forward pass and mixes the value of having memory with the value of learning through it. Holding the forward computation fixed eliminates this confound, but a second ambiguity remains: the standard stop-gradient cut at stored history removes two mechanisms at once. One updates the network parameters that determine what to write into memory (\emph{representation credit}), while the other travels back through the physical states that earlier actions produced and that later entered memory (\emph{closed-loop credit}). Static gradient norms, the usual diagnostic for truncation cuts, help little here because they capture the learning signal at only a single moment of training.

We therefore construct forward-matched ablation variants, which we call \emph{arms}, that keep every forward rollout identical and remove one backward route at a time (Section~\ref{sec:routes}). Using quadrotor tracking as our primary testbed, where noisy or hidden velocity creates a concrete demand for memory, we vary velocity observability and weaken multi-step physical credit by decaying the gradient along each state-to-state edge. A learned autoregressive model of vessel motion \citep{gan2026sketch}, whose state is fully observed, provides a full-state counterpart, while two additional underactuated systems (a fixed-wing aircraft and a cart-pole) test how far the separation extends beyond the quadrotor.

Three findings stand out. First, credit through memory improved learning only when velocity was noisy or hidden; with exact velocity, cutting memory gradients incurred almost no truncation cost. Second, when velocity was hidden in the quadrotor, a mild decay of state-to-state physical gradients ($\gamma = 0.8$) eliminated the truncation cost of removing closed-loop credit while representation credit remained essential, a separation that held in 14 of 16 new seeds. Third, in a fixed-wing aircraft closed-loop credit could be replaced by representation credit, leaving no cost for state-edge decay to separate, while in a cart-pole training with full state-edge credit was too unstable to compare the routes. Finally, static gradient magnitudes did not predict which cut cost more in either the quadrotor or the vessel model.

The paper thus contributes a split of credit through memory into two routes with arms that keep every forward rollout fixed (Section~\ref{sec:routes}), measurements of when each route matters and when one can replace the other (Section~\ref{sec:results}), and evidence that gradient norms predict neither (Section~\ref{sec:grad_results}).

\section{Related work}
\label{sec:related}

\textbf{Truncating credit through memory.} Truncated backpropagation through time limits how far gradients travel through a recurrent network \citep{williams1990efficient}, and the resulting bias can be removed \citep{tallec2017unbiasing} or controlled by choosing the truncation length adaptively \citep{aicher2020adaptive}. Transformer-XL caches the hidden states of earlier segments and stops gradients at them \citep{dai2019transformerxl}, and GTrXL carries this memory architecture into reinforcement learning \citep{parisotto2020gtrxl}, while sparse attentive backtracking sends credit selectively to a small subset of past states \citep{ke2018sab}. Closest to our question, \citet{ni2023memory} separate context memory from credit assignment in model-free reinforcement learning and show that Transformers extend the first while leaving long-horizon credit assignment as hard as before. Yet in all of these settings credit through memory stays inside the network because the environment is not differentiated. Our setting adds a differentiable environment, allowing credit that reaches a stored observation to leave the network and continue backward through the physics.

\textbf{Analytic policy gradients.} Whether gradients from a differentiable simulator aid optimization depends on the smoothness, stiffness, and chaos of the system \citep{suh2022gradients}. \citet{metz2021gradients} trace exploding gradients to the spectrum of the dynamics' Jacobian and find that only a narrow band of truncation lengths trains well. Several methods therefore limit or reshape the physical backward path: short-horizon and adaptive-horizon actor-critics shorten the rollout \citep{xu2022shac,georgiev2024ahac}, Soft Analytic Policy Optimization stabilizes learning in differentiable multiphysics simulation \citep{xing2025sapo}, \citet{zhang2025agile} decay gradients along state edges, and \citet{schnell2024stabilizing} modify the backward pass for physics directly, noting that the resulting update is no longer a gradient field. Quadrotor controllers have been trained this way from states and from visual features \citep{wiedemann2023apg,heeg2025learning}. Some APG controllers with memory also stop gradients at the policy input: \citet{nachkov2025controllers} backpropagate through the hidden state of a recurrent driving controller while detaching its observation and simulator state, and \citet{you2025accelerating} detach rendered observations; both correspond to our no-observation-gradient control. With learned Transformer world models in place of a simulator, circuitous gradient paths can similarly hurt long-horizon policy gradients \citep{ma2024transformer}. Whereas those works decide how far credit travels through the physics, we ask what happens to the part of it that passes through memory.

\begin{figure}[t]
\centering
\begin{tikzpicture}[
    font=\sffamily\footnotesize,
    >={Stealth[length=4.2pt,width=3.4pt]},
    var/.style={circle, draw=black!65, line width=0.55pt, fill=white, minimum size=7.4mm, inner sep=0pt, font=\small},
    store/.style={var, fill=storefill},
    par/.style={rectangle, rounded corners=1.2pt, draw=black!55, line width=0.5pt, fill=black!6, minimum width=5.2mm, minimum height=4.6mm, inner sep=0pt, font=\small},
    fwd/.style={->, draw=black!32, line width=0.55pt},
    cred/.style={->, draw=credgray, line width=1.25pt},
    rep/.style={->, draw=repblue, line width=1.6pt},
    loop/.style={->, draw=loopverm, line width=1.6pt},
    lab/.style={font=\sffamily\scriptsize, text=black!55},
]
\def\xa{0}\def\xb{2.35}\def\xc{5.55}\def\xd{7.55}
\def\ys{0}\def\yo{1.3}\def\yn{2.6}\def\ya{3.9}
\node[var]   (sjm) at (\xa,\ys) {$s_{j-1}$};
\node[var]   (ajm) at (\xa,\ya) {$a_{j-1}$};
\node[var]   (sj)  at (\xb,\ys) {$s_j$};
\node[var]   (oj)  at (\xb,\yo) {$o_j$};
\node[store] (nj)  at (\xb,\yn) {$n_j$};
\node[var]   (st)  at (\xc,\ys) {$s_t$};
\node[var]   (ot)  at (\xc,\yo) {$o_t$};
\node[var]   (nt)  at (\xc,\yn) {$n_t$};
\node[var]   (at)  at (\xc,\ya) {$a_t$};
\node[var]   (stp) at (\xd,\ys) {$s_{t+1}$};
\node[font=\small] (loss) at (\xd+0.75,\ys) {$\ell$};
\node[par]   (thA) at (\xa-0.95,\ya) {$\theta$};
\node[par]   (thN) at (\xb-1.25,\yn+0.45) {$\theta$};
\draw[fwd] (sjm) -- node[lab, below=0.5pt, pos=0.5] {state edge} (sj);
\draw[fwd] (ajm) -- (sj);
\draw[fwd] (sj) -- (oj);
\draw[fwd] (oj) -- (nj);
\draw[fwd, densely dotted] (sj) -- (st);
\draw[fwd] (st) -- (ot);
\draw[fwd] (ot) -- (nt);
\draw[fwd] (nt) -- (at);
\draw[fwd] (st) -- (stp);
\draw[fwd] (at) -- (stp);
\draw[fwd, dashed, draw=black!45] (nj) -- node[lab, above=1pt, pos=0.5] {memory read} (nt);
\draw[draw=black!32, line width=0.55pt] (stp) -- (loss);
\draw[cred] (stp.north) to[out=95, in=0] node[lab, right=2pt, pos=0.45, align=left, text=credgray] {credit from\\a later loss} (at.east);
\draw[cred] (at.south east) to[bend left=45] (nt.north east);
\draw[cred] (nt.north west) to[out=150, in=30] node[lab, above=0pt, pos=0.5, text=credgray, align=center] {credit reaching\\the stored $n_j$} (nj.north east);
\draw[rep] (nj.west) to[out=180, in=-20] (thN.east);
\node[font=\sffamily\scriptsize\bfseries, text=repblue, align=left, anchor=south west] at (\xb-0.98,\yn+0.62) {representation\\credit};
\draw[loop] (nj.south east) to[bend left=38] (oj.north east);
\draw[loop] (oj.south east) to[bend left=38] (sj.north east);
\draw[loop] (sj.west) to[out=165, in=-95] (ajm.south);
\draw[loop] (ajm.west) to[out=180, in=0] (thA.east);
\draw[loop, dashed] (sj.south) to[out=-115, in=-65, looseness=1.05] node[pos=0.5, above=-1pt, font=\small, text=loopverm] {$\times\gamma$} (sjm.south);
\node[font=\sffamily\scriptsize\bfseries, text=loopverm, align=left, anchor=west] at (\xb+0.62,\yo) {closed-loop\\credit};
\foreach \x/\s in {\xa/{step $j-1$}, \xb/{step $j$}, \xc/{step $t$}, \xd/{step $t+1$}}
    \node[lab, text=black!70] at (\x,-1.25) {\s};
\node[font=\sffamily\bfseries\small, anchor=north west] at (-1.55,4.55) {a};
\begin{scope}[xshift=10.15cm, yshift=0.35cm]
\node[font=\sffamily\bfseries\small, anchor=north west] at (-0.3,4.2) {b};
\node[font=\sffamily\scriptsize\bfseries, text=repblue, align=center] at (3.05,3.55) {repr.};
\node[font=\sffamily\scriptsize\bfseries, text=loopverm, align=center] at (3.85,3.55) {closed\\loop};
\foreach \i/\name/\r/\l in {0/{full policy}/1/1, 1/{no representation credit}/0/1, 2/{no closed-loop credit}/1/0, 3/{no credit through memory}/0/0} {
    \pgfmathsetmacro{\yy}{2.85-0.75*\i}
    \node[anchor=west, font=\sffamily\scriptsize, text width=2.55cm, align=left] at (-0.3,\yy) {\name};
    \ifnum\r=1 \fill[repblue] (3.05,\yy) circle (1.15mm); \else \draw[repblue, line width=0.9pt] (3.05,\yy) circle (1.05mm); \fi
    \ifnum\l=1 \fill[loopverm] (3.85,\yy) circle (1.15mm); \else \draw[loopverm, line width=0.9pt] (3.85,\yy) circle (1.05mm); \fi
}
\node[anchor=north west, font=\sffamily\scriptsize, text=black!60, text width=4.4cm, align=left] at (-0.3,0.25) {Every arm produces the same rollout;\\only the backward route changes.};
\end{scope}
\end{tikzpicture}
\caption{Two routes for credit through memory. \textbf{(a)} A step $j$ of a rollout whose stored representation $n_j$ is read at a later step $t$. Credit that reaches $n_j$ can end at the parameters $\theta$ inside its computation (representation credit, blue) or leave through the observation $o_j$ into the state $s_j$ and continue to earlier actions (closed-loop credit, vermilion). The state-edge decay $\gamma$ scales every state-to-state edge. \textbf{(b)} Each ablation arm keeps (filled) or removes (open) each route. All arms produce the same rollout and differ only in where the learning signal goes.}
\label{fig:concept}
\end{figure}

\textbf{What memory stores.} What a recurrent network stores is shaped by the backward learning signal it receives. The e-prop algorithm, for instance, factorizes the gradient of a recurrent network into local eligibility traces and top-down learning signals \citep{bellec2020solution}, and \citet{leeftink2026neural} interpret the hidden states of recurrent policies as Pontryagin costates, structuring them by supervising the policy with costate targets. Rather than altering the learning signal or adding supervision, we decompose the natural end-to-end gradient reaching memory and measure what each route contributes.

\section{Two routes for credit through memory}
\label{sec:routes}

Credit for a later tracking error can reach a policy's memory along two routes. Here \emph{credit} denotes the part of the loss gradient that reaches a given quantity, and \emph{memory} denotes the representations a policy writes during an episode and reads at later steps; the optimizer's state, for example, is not part of it. We motivate the split with a physical control example, formalize it for a minimal scalar system, and then define it for the Transformer policies we train (Figure~\ref{fig:concept}).

\subsection{A running example}
\label{sec:running_example}

Consider a quadrotor that must track a reference trajectory without observing its own velocity. Its policy sees where it is relative to the reference, its attitude, and its angular velocity, so it has to infer linear velocity from the history of positions it has stored. When a tracking error at a later step depends on that inference, gradients flowing back into memory can improve the policy in two ways. First, they can change how the network turns stored positions into a velocity estimate, which we call \emph{representation credit}. Alternatively, they can exit through past observations to reach the earlier thrust and body-rate commands that moved the quadrotor to those positions, teaching the policy how its own actions show up in what it later stores; we call this \emph{closed-loop credit}. When velocity is hidden, closed-loop credit can leave memory only through the observed inputs that remain, such as relative position and attitude.

\subsection{A minimal example}
\label{sec:minimal_example}

Mathematically, the two routes correspond to the two terms of a product rule. Take a scalar state that integrates the action, $s_j = s_{j-1} + a_{j-1}$, a policy that stores $n_j = w\,s_j$, and a later action that reads the store, $a_t = u\,n_j$ with $t > j$. Because earlier actions come from the same policy parameter $w$, the state $s_j$ also depends on $w$, and the derivative that passes through the store splits into
\begin{equation}
\label{eq:minimal}
\frac{\partial a_t}{\partial n_j}\,\frac{\mathrm{d}n_j}{\mathrm{d}w}
= \underbrace{u\,s_j}_{\text{representation}}
+ \underbrace{u\,w\,\frac{\mathrm{d}s_j}{\mathrm{d}w}}_{\text{closed-loop}}.
\end{equation}
The first term ends at $w$ inside the computation of the store. The second leaves the store through the physical state $s_j$, crosses the dynamics, and reaches $w$ again through the earlier actions that produced $s_j$; it exists only because the gradient passes through the environment. Writing down Equation~\eqref{eq:minimal} is straightforward, but knowing what each term contributes to learning is another matter: in a full policy the second term can itself pass through earlier stores, causing the two routes to mix.

\begin{figure}[t]
\small
\noindent\rule{\textwidth}{0.8pt}\vspace{-4pt}
\textbf{Algorithm 1} One step $j$ of a rollout in each ablation arm. \texttt{n} holds the normalized inputs to attention of every layer, and \texttt{history} the stored copies of earlier steps. All arms compute the same forward values; they differ only in which derivatives the stored copy passes back.\vspace{-4pt}\\
\noindent\rule{\textwidth}{0.5pt}
\begin{verbatim}
out_j, n_live = policy(o_j, history)              # the policy acts at step j
if arm == "no representation credit":
    _, n_store = policy[detach(theta)](o_j, history)  # detached parameters
elif arm == "no closed-loop credit":
    _, n_store = policy(detach(o_j), history)         # detached observation
elif arm == "no credit through memory":
    n_store = detach(n_live)
else:                                             # full policy
    n_store = n_live
history.append(n_store)   # later steps project it with the current W_K, W_V
\end{verbatim}
\vspace{-8pt}\noindent\rule{\textwidth}{0.8pt}
\end{figure}

\subsection{Definitions}
\label{sec:definitions}

We now formalize this decomposition for a causal Transformer policy trained through a differentiable simulator. The state evolves as $s_{j+1} = f(s_j, a_j)$, and the policy observes $o_j = h(s_j, \xi_j)$, where $\xi_j$ adds noise or masks inputs. In each attention layer $\ell$, step $j$ stores $n_j^\ell = g_\theta^\ell(o_j, n_{<j})$, the normalized input to attention before the key and value projections, and a later step $t > j$ reads $n_j^\ell$ by projecting it with the current key and value weights.

Once a backward path enters a stored representation $n_j$ from a later step, it can end in only two ways: either it terminates at the parameters $\theta$ inside the computation of $n_j$, or it leaves that computation through the observation $o_j$, continues into the state $s_j$, and runs back through $f$ to earlier actions. Pure paths of the first kind form $P_{\mathrm{rep}}$ and pure paths of the second kind form $P_{\mathrm{loop}}$. Some composite paths do both, leaving one store through an observation and later entering an earlier store where they end at $\theta$; these mixed paths form $P_{\mathrm{mix}}$. Writing $g_0$ for the contribution of all paths that never enter a store, the full gradient is $g_{\mathrm{full}} = g_0 + P_{\mathrm{rep}} + P_{\mathrm{loop}} + P_{\mathrm{mix}}$.

Each ablation arm redirects this backward signal while leaving the stored values unchanged (Algorithm~1). At every step we compute the policy's representations twice: a live copy $n_{\mathrm{live}}$ that the current step uses, and an identical stored copy $n_{\mathrm{store}}$ that later steps read. Applying the stop-gradient operator $\sg$ (the identity on values and zero on derivatives) only inside $n_{\mathrm{store}}$ guarantees that every forward rollout is identical across arms. Removing representation credit computes the stored copy with detached parameters, $g_{\sg(\theta)}(o_j, \tilde{n}_{<j})$; removing closed-loop credit computes it from a detached observation, $g_\theta(\sg(o_j), \tilde{n}_{<j})$; and removing all credit through memory stores $\sg(n_{\mathrm{live}})$. Because $\tilde{n}_{<j}$ are themselves stored copies, a backward path that continues into earlier stores meets the same restriction recursively. In every arm the key and value weights still receive gradients from historical positions, as in Transformer-XL \citep{dai2019transformerxl}. The four arm gradients satisfy
\begin{equation}
\label{eq:four_arms}
g_{\mathrm{full}} = g_0 + P_{\mathrm{rep}} + P_{\mathrm{loop}} + P_{\mathrm{mix}}, \quad
g_{\text{no-rep}} = g_0 + P_{\mathrm{loop}}, \quad
g_{\text{no-loop}} = g_0 + P_{\mathrm{rep}}, \quad
g_{\text{no-mem}} = g_0,
\end{equation}
so that $g_{\mathrm{full}} - g_{\text{no-rep}} - g_{\text{no-loop}} + g_{\text{no-mem}} = P_{\mathrm{mix}}$. Removing either route individually thus also removes the mixed paths that the two routes share.

\begin{table}[t]
\caption{Ablation arms. Each removes one class of backward paths, and only the short-memory controls change the forward computation.}
\label{tab:arms}
\begin{center}
\small
\begin{tabular}{llP{0.25\linewidth}l}
\toprule
Arm & Credit removed & Forward computation & Used in \\
\midrule
full policy & none & reference & all sections \\
no representation credit & $P_{\mathrm{rep}}$ and $P_{\mathrm{mix}}$ & identical & \ref{sec:when_matters}--\ref{sec:grad_results} \\
no closed-loop credit & $P_{\mathrm{loop}}$ and $P_{\mathrm{mix}}$ & identical & \ref{sec:when_matters}--\ref{sec:grad_results} \\
no credit through memory & all paths through memory & identical & \ref{sec:when_matters}, \ref{sec:other_systems}, \ref{sec:grad_results} \\
no observation gradient & every path through an observation & identical & \ref{sec:when_matters} \\
short memory ($w = 1, 2$) & none (memory restricted) & attention window $w$ & \ref{sec:when_matters} \\
\bottomrule
\end{tabular}
\end{center}
\end{table}

\subsection{Arms and state-edge decay}
\label{sec:controls_decay}

Removing one class of paths at a time gives three memory-cut arms, and two further configurations act as controls (Table~\ref{tab:arms}). The no-observation-gradient control stops gradients at every observation, the current one included; it keeps internal memory paths that end at $\theta$ and removes all credit flowing from observations back into the physics. The short-memory controls restrict attention to the current step ($w = 1$) or to the current and previous steps ($w = 2$). Because they are the only variants that change the forward computation, we use them to measure what reading memory itself is worth.

To weaken credit through physical states without changing the forward rollout, we decay the gradient along every state-to-state edge \citep{zhang2025agile}. At step $j$, the simulator receives $\tilde{s}_j = \gamma\,s_j + (1 - \gamma)\,\sg(s_j)$ in place of $s_j$. Its forward output is unchanged, while the derivative across each state-to-state edge is multiplied by $\gamma$ and credit crossing $k$ such edges is scaled by $\gamma^k$. Setting $\gamma = 1$ preserves exact gradients, while $\gamma = 0$ retains only single-step action-to-state edges. Observations read the undecayed state $s_j$, so observation edges keep their full gradient. We refer to $\gamma$ as the \emph{state-edge decay}.

\section{Experimental setup}
\label{sec:setup}

Comparing ablation arms is informative only if they differ solely in their backward graphs. Within each seed, all arms share the parameter initialization, the sequence of training batches and observation noise, the optimizer settings, and the evaluation panel. Before training, we checked numerically that all forward-matched arms produce identical rollouts and the intended gradients of Equation~\eqref{eq:four_arms} (Appendix~\ref{app:implementation}).

\textbf{Quadrotor.} We use the 12-state quadrotor dynamics of \citet{wiedemann2023apg} with a refined integrator (Appendix~\ref{app:systems}). The policy outputs collective thrust and three body rates every $0.05\text{ s}$ over a 32-step episode to track a reference whose velocity varies sinusoidally on each axis (amplitudes $0.1\text{--}0.3\text{ m/s}$). The controller is a two-layer causal Transformer (width 64, 4 heads) attending to all earlier steps of the episode. We train each arm from initialization for 1000 AdamW updates and measure its final error as the time-averaged squared position error on a fixed panel of 256 trajectories across four velocity-observability conditions: \emph{exact}, Gaussian noise of standard deviation $0.05$ or $0.20\text{ m/s}$ on every velocity input, and \emph{hidden} (all velocity inputs set to zero). We train 16 initial seeds per condition and 16 additional seeds for the replication in Section~\ref{sec:separation}.

\textbf{Vessel.} As a full-state counterpart whose state is observed without noise, we adapt a learned model of commercial vessel motion on real ship tracks \citep{gan2026sketch}, treating the autoregressive predictor itself as the policy. Over a one-day rollout in five-minute steps (288 steps), its output at each step sets the next velocity, spherical kinematics integrate velocity into position, and the model reads both back without noise; credit reaching its memory thus takes the same two routes as in the quadrotor. Starting from a backbone pretrained with teacher forcing, we add a zero-initialized linear correction head to constant-velocity extrapolation and train with full physical credit across 16 sampling streams, evaluating on held-out track segments (Appendix~\ref{app:systems}). Because gradient clipping was active on $10\text{--}14\%$ of updates depending on the arm, we treat the vessel results as descriptive observations (Appendix~\ref{app:vessel}).

\textbf{Truncation costs and statistics.} For each seed we divide the final error $E_{\mathrm{arm}}$ of an arm by that of the full policy trained with the same seed, observability condition, and $\gamma$. We call the relative excess error, $c = E_{\mathrm{arm}}/E_{\mathrm{full}} - 1$, the arm's \emph{truncation cost} and average it over seeds: a truncation cost of $20\%$ means that training without that route ends $20\%$ worse than the full policy, directly measuring how much the route helps learning. Differences between truncation costs are given in percentage points. Following \citet{agarwal2021deep}, seeds (or vessel streams) are the unit of analysis, and we report means with $95\%$ percentile bootstrap intervals rather than significance tests; an interval that includes zero leaves the direction of an effect open rather than showing no effect. For the main comparisons we fixed the summary statistic and interval width before the corresponding runs, widening the interval where several predictions shared the same runs (Appendix~\ref{app:predictions}). Because a single evaluation at update 1000 can catch a run during a transient error spike, Appendices~\ref{app:quadrotor} and~\ref{app:other_systems} also give the main comparisons under five alternative measures of the final error.

\section{Results}
\label{sec:results}

\subsection{Credit through memory matters only when velocity is unreliable}
\label{sec:when_matters}

\begin{figure}[t]
\centering
\includegraphics[width=\textwidth]{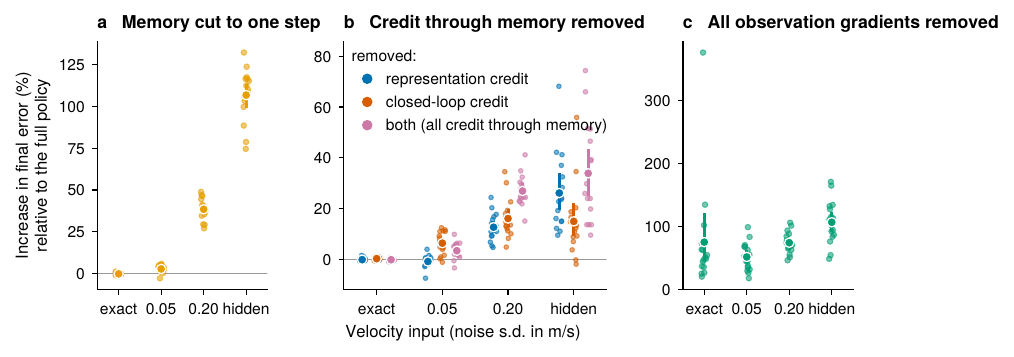}
\caption{Credit through memory matters only when velocity is unreliable. Quadrotor, $\gamma = 1$. Truncation cost is the increase in final error relative to the full policy trained with the same seed. \textbf{(a)} Memory restricted to the current step. \textbf{(b)} Removing representation credit, closed-loop credit, or both. \textbf{(c)} Stopping gradients at every observation. Small points are seeds (16 per condition), and large points and bars are means with $95\%$ intervals. Vertical scales differ between panels.}
\label{fig:observability}
\end{figure}

Credit through memory improved quadrotor learning only when the velocity input was unreliable (Figure~\ref{fig:observability}b; Table~\ref{tab:C1}). With exact velocity, no cut changed the final tracking error by more than $0.3\%$ on average. Under $0.20\text{ m/s}$ velocity noise or hidden velocity, by contrast, removing one route or both raised the final error by $13\%$ to $34\%$ on average, and removing all credit through memory raised it in every seed of both conditions.

Forward memory availability followed the same pattern (Figure~\ref{fig:observability}a). Restricting attention to the current step ($w = 1$) cost nothing with exact velocity ($-0.3\%$) and roughly doubled tracking error with velocity hidden ($+107\%$). Thus, when the policy must estimate velocity from history, both reading memory and backpropagating through it pay off, though forward memory access carries the larger share: with velocity hidden, keeping forward memory while cutting all credit through it ended $34\%$ worse than the full policy, against $107\%$ worse for a policy without memory.

Stopping gradients at every observation cost far more across all four observability regimes, raising final error by $51\%$ to $107\%$ on average across the four conditions, in every seed, and even with exact velocity (Figure~\ref{fig:observability}c). The size of this cost surprised us. A likely reason is that, without these gradients, optimization no longer sees how the policy's reaction to a state deviation changes the states that follow. Since this control also removes closed-loop credit through memory (which on its own cost at most $16\%$), most of its cost appears to come from direct observation paths that bypass memory.

The fully observed vessel model behaved like the quadrotor with exact velocity: no memory cut changed its final error by more than about $1.5\%$ (the largest upper bound of the $95\%$ intervals across 16 streams), leaving the order of the two routes unresolved (Table~\ref{tab:2}; Appendix~\ref{app:vessel}). Removing closed-loop credit, however, made vessel training numerically more fragile. The two arms without it produced 243 numerical events (single-precision gradient-norm overflows or non-finite gradients), against 13 for the two arms that kept it and none for the control without observation gradients. We first noticed this pattern in six streams, and among ten further streams it held in all nine that had any events (Appendix~\ref{app:vessel}); the quadrotor runs had none, and Section~\ref{sec:discussion} returns to these events.

\subsection{Observation noise raises the truncation cost of both routes}
\label{sec:noise_both}

Because standard cache truncation removes both routes at once, Figure~\ref{fig:observability}b leaves open which route carries the learning benefit. We had expected a division of labor: observation noise would primarily raise the cost of removing representation credit (since noisy velocity is what memory has to summarize), while multi-step physical credit would primarily feed closed-loop credit. The first half of that expectation did not hold. Going from exact to noisy velocity ($0.20\text{ m/s}$) at $\gamma = 1$ raised the truncation cost of both routes from near zero to similar levels: $13\%$ for representation credit and $16\%$ for closed-loop credit (Figure~\ref{fig:observability}b; Appendix~\ref{app:predictions} gives the interval for their difference). Noise, in other words, made both routes more valuable.

When velocity was hidden at $\gamma = 1$, each route again contributed substantially on its own, though which cut cost more varied between the two 16-seed sets: removing representation credit cost more in the initial 16 seeds ($26\%$ vs.\ $15\%$, with the same order in 13 of 16 seeds), while in the second set of 16 seeds (Section~\ref{sec:setup}) the two mean truncation costs were nearly equal ($20\%$ and $19\%$; Table~\ref{tab:C3}). At low velocity noise ($0.05\text{ m/s}$), only closed-loop credit had a measurable truncation cost ($6.3\%$, against $-0.9\%$ for representation credit; Appendix~\ref{app:quadrotor}).

\subsection{Mild state-edge decay separates representation credit from closed-loop credit}
\label{sec:separation}

Although both routes contribute when velocity is hidden or noisy at $\gamma = 1$, they reach back through the physics to different extents: closed-loop credit must travel from a stored observation back through the chain of physical states that produced it, whereas representation credit ends inside the network that wrote the memory. This structural difference suggests a test for the second half of our division-of-labor hypothesis: weakening credit through physical states via state-edge decay ($\gamma < 1$) should reduce the truncation cost of the closed-loop route before that of representation credit.

\begin{figure}[t]
\centering
\includegraphics[width=\textwidth]{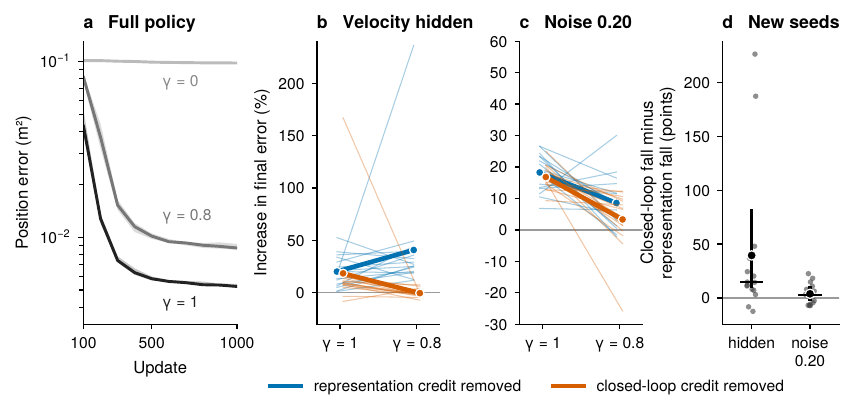}
\caption{Mild state-edge decay removes the contribution of closed-loop credit. Quadrotor. \textbf{(a)} Error of the full policy during training at $\gamma = 1$, $0.8$, and $0$, velocity hidden (16 seeds). \textbf{(b, c)} Truncation cost of each route, the increase in final error when it is removed, at $\gamma = 1$ and $0.8$ in the 16 new seeds, with velocity hidden \textbf{(b)} or noisy \textbf{(c)}. Thin lines are seeds and thick lines means. \textbf{(d)} How much more truncation cost fell for closed-loop credit than for representation credit from $\gamma = 1$ to $0.8$, per seed; positive values follow the prediction. Black points and bars are means with $97.5\%$ intervals, widened because two predictions share these runs (Appendix~\ref{app:systems}), and horizontal ticks are medians. With velocity hidden, closed-loop credit lost its value at $\gamma = 0.8$ while representation credit kept it.}
\label{fig:separation}
\end{figure}

Our first test, conducted under $0.20\text{ m/s}$ velocity noise at the strongest decay of $\gamma = 0$, did not support the prediction, because complete state-edge truncation stalled training and left the comparison uninformative: without state-to-state credit, training stalled after about 100 updates at $16$ times the error of the $\gamma = 1$ full policy ($18$ times with velocity hidden; Figure~\ref{fig:separation}a), and at $0.20\text{ m/s}$ noise no single cut cost more than $1.4\%$ of the $\gamma = 1$ full-policy error (Appendix~\ref{app:predictions}). A mild decay of $\gamma = 0.8$, by contrast, slowed training without stalling it, leaving the full policy at $1.6$ to $1.9$ times its $\gamma = 1$ error. On the initial 16 seeds with velocity hidden, lowering $\gamma$ from $1$ to $0.8$ eliminated the truncation cost of closed-loop credit (which fell from $15\%$ to $-1.4\%$), while the truncation cost of representation credit rose from $26\%$ to $34\%$ (Table~\ref{tab:C2}).

Because we selected $\gamma = 0.8$ only after examining the original 16 seeds, we trained a second set of 16 seeds, with the prediction fixed before training them: from $\gamma = 1$ to $\gamma = 0.8$, the truncation cost of closed-loop credit should fall by more than that of representation credit. The new seeds bore this out (Figure~\ref{fig:separation}b, d; Table~\ref{tab:C3}). With velocity hidden, removing closed-loop credit again incurred no cost at $\gamma = 0.8$ ($-0.4\%$ mean, $-1.3\%$ median), while representation credit mattered in every seed, with a median truncation cost of $26\%$. Across the 16 new seeds, the truncation cost of closed-loop credit fell by more than that of representation credit in 14 of the 16 seeds, by a median of 15 percentage points (Figure~\ref{fig:separation}d); the mean difference of 39 points ($97.5\%$ interval $[10, 81]$) is larger because two seeds spiked in error near update 1000. Measuring the costs against the $\gamma = 1$ full-policy error, or under the five alternative final-error measures of Appendix~\ref{app:quadrotor}, gives the same picture. Weaker optimization does not explain the separation either: at $\gamma = 0.8$ training did not stall, and representation credit kept its value in the same runs.

With noisy velocity ($0.20\text{ m/s}$), however, $\gamma = 0.8$ reduced the truncation cost of both routes simultaneously (closed-loop cost falling from $17\%$ to $3.3\%$ and representation cost from $18\%$ to $8.6\%$; Figure~\ref{fig:separation}c), leaving only a small, unresolved difference of $3.8$ points (Figure~\ref{fig:separation}d). The separation between routes is therefore clearest when velocity is missing altogether. One possible reason is that a policy without any direct velocity signal must construct a velocity estimate in memory, and representation credit is what shapes that estimate; closed-loop credit, by contrast, would rely on multi-step chains through the physics long enough for a mild decay to weaken them. We have not tested this explanation directly.

\subsection{The two routes also differ in how far back and when they act}
\label{sec:span_dynamics}

\begin{figure}[t]
\centering
\includegraphics[width=\textwidth]{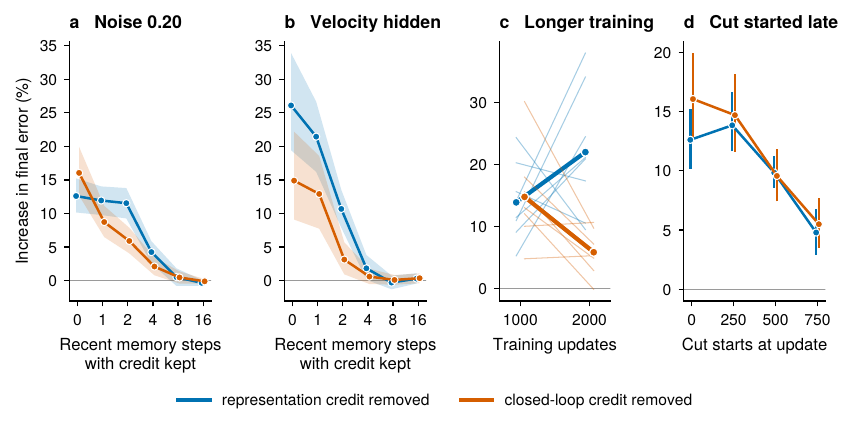}
\caption{Further differences between the two routes. Quadrotor, noise $0.20\text{ m/s}$ unless stated. \textbf{(a, b)} Truncation cost when credit is kept for the $s$ most recent stored steps and removed for older ones, with noisy \textbf{(a)} or hidden \textbf{(b)} velocity, where $s = 0$ is the full cut. \textbf{(c)} Truncation cost after 1000 and 2000 updates in 8 seeds, with thin lines for seeds. \textbf{(d)} Truncation cost when the cut starts at a later update, with the optimizer state carried over. Bands and bars are $95\%$ intervals over 16 seeds unless stated.}
\label{fig:differences}
\end{figure}

To understand why the two routes behave differently, we ran three follow-up experiments on the original quadrotor seeds, which we did not repeat on the second seed set (Figure~\ref{fig:differences}; Appendix~\ref{app:further} gives all values).

\textbf{Temporal span across stored history.} Closed-loop credit acted mainly through the most recent stored steps, while representation credit drew on slightly older steps. When we kept credit through the $s$ most recent stored steps and removed it from older ones under $0.20\text{ m/s}$ noise (Figure~\ref{fig:differences}a), keeping a single step ($s = 1$) eliminated about half of the closed-loop truncation cost and almost none of the representation cost, and beyond eight steps neither cut had a measurable cost. This is the reverse of what we expected, since representation credit had seemed the more local of the two routes. When velocity was hidden, the point estimates pointed the same way, though the contrast was not established (Figure~\ref{fig:differences}b).

\textbf{Training length.} Longer training reversed the order of the two truncation costs (Figure~\ref{fig:differences}c). Across 8 seeds trained twice as long (2000 updates at noise $0.20\text{ m/s}$), the closed-loop truncation cost fell from $15\%$ to $5.8\%$ while the representation cost rose from $14\%$ to $22\%$ (the interval of the fall excludes zero, while that of the rise does not; Appendix~\ref{app:further}). Closed-loop credit may thus mainly speed up early learning while representation credit governs what is eventually learned, so our comparisons at 1000 updates characterize a fixed training budget.

\textbf{Late-stage truncation.} Credit through memory also kept mattering late in training (Figure~\ref{fig:differences}d). Switching from the full policy to a cut at update 250 cost about as much as cutting from the start ($14\%$ and $15\%$ for representation and closed-loop credit, against $13\%$ and $16\%$), and even switching at update 750 still cost about $5\%$. These switches carry over the AdamW optimizer state, because resetting the optimizer state alone without a cut lowered the error by $7.2\%$ and $5.5\%$ at updates 250 and 500 and would have confounded the comparison (Appendix~\ref{app:further}).\footnote{Running the switch in the opposite direction suggests that part of what a closed-loop cut leaves behind may be held in the optimizer state rather than in the weights: when the cut was lifted at update 500, a cost of $6.3\%$ remained while the optimizer state was kept and mostly disappeared after a reset (Appendix~\ref{app:further}).}

\subsection{In a fixed-wing aircraft, closed-loop credit could be replaced}
\label{sec:other_systems}

To see whether the separation holds beyond the quadrotor, we tested two further underactuated systems with velocity hidden, which are the only additional systems we examined: a fixed-wing aircraft flying toward a target \citep{wiedemann2023apg} and a frictionless cart-pole written for this study that balances its pole while tracking a reference, both using the quadrotor policy architecture and arms (Appendix~\ref{app:other_systems}).

In the fixed-wing aircraft, representation credit could stand in for closed-loop credit on its own (Table~\ref{tab:2}). At $\gamma = 1$, removing closed-loop credit alone left the tracking error almost unchanged (median $0.3\%$), while removing both routes raised error in 11 of 16 seeds (median $5.8\%$) and removing representation credit alone lay in between (median $4.9\%$). All three median costs were small, and the closed-loop cost stayed within $0.3\%$ of zero under every final-error measure we tried (Appendix~\ref{app:other_systems}). Short pilot runs had already hinted that removing closed-loop credit cost very little here, and accordingly the 16-seed decay comparison found no separation: with $\gamma = 0.9$, the truncation cost of closed-loop credit fell by more than that of representation credit in only 7 of the 16 seeds (Appendix~\ref{app:other_systems}).

In the cart-pole, by contrast, training with full state-edge credit was too unstable to compare the routes reliably. Its upright equilibrium is open-loop unstable, and at $\gamma = 1$, 23 of the 64 runs relapsed, rising to at least twice their lowest error by update 1000. Single-checkpoint costs at update 1000 thus depended on whether a run happened to be relapsing at that moment, and alternative evaluation windows flipped even the median ordering of the routes; we therefore draw no conclusion about the routes from this system, and the planned decay comparison did not support the separation either (Appendix~\ref{app:other_systems}).

\begin{table}[t]
\caption{Truncation cost of each route in three systems, $\gamma = 1$: medians over seeds of the increase in final error when the route is removed, with the number of seeds in which the error increased. We use medians because a few fixed-wing seeds had large costs. Means with intervals are in Section~\ref{sec:when_matters} and Appendices~\ref{app:quadrotor}, \ref{app:other_systems}, and~\ref{app:vessel}, and the cart-pole, whose training was too unstable to compare the routes, is reported in Appendix~\ref{app:other_systems}.}
\label{tab:2}
\begin{center}
\small
\begin{tabular}{llccc}
\toprule
 & & \multicolumn{3}{c}{Credit removed} \\
\cmidrule(lr){3-5}
System & Velocity & both routes & closed-loop only & representation only \\
\midrule
Quadrotor, 16 seeds & hidden & 28\% (16/16) & 13\% (14/16) & 23\% (16/16) \\
Fixed-wing, 16 seeds & hidden & 5.8\% (11/16) & 0.3\% (10/16) & 4.9\% (11/16) \\
Vessel, 16 streams & observed & $-0.5\%$ (6/16) & $-0.4\%$ (6/16) & 1.0\% (12/16) \\
\bottomrule
\end{tabular}
\end{center}
\end{table}

Thus the fixed-wing aircraft lacks what the separation needs: where removing closed-loop credit costs nothing to begin with, decaying state-edge credit cannot lower that cost further, so the decay has nothing to act on. Its precondition, a closed-loop route that the representation route cannot replace on its own, held in the quadrotor with hidden velocity, where each route had value of its own.

\subsection{Gradient magnitude does not predict truncation cost}
\label{sec:grad_results}

\begin{figure}[t]
\centering
\includegraphics[width=0.80\textwidth]{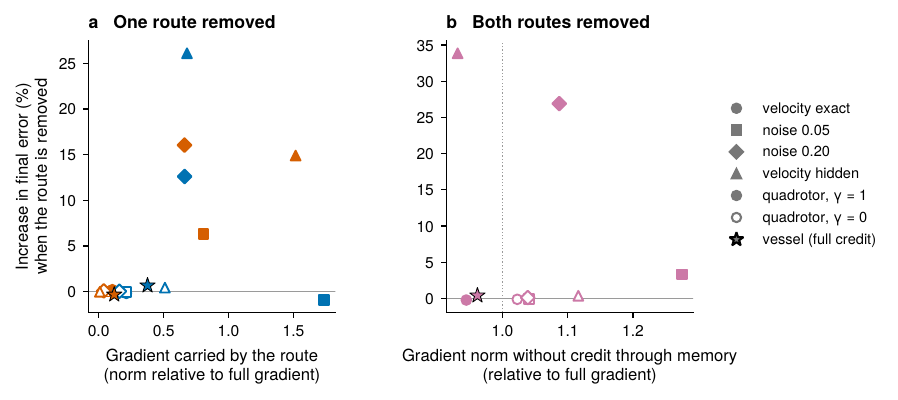}
\caption{Gradient magnitude does not predict truncation cost. \textbf{(a)} Norm of one route's contribution, relative to the full gradient, against the truncation cost of removing that route, for the quadrotor at $\gamma = 1$ (filled) and $\gamma = 0$ (open) under four velocity conditions, and the vessel model (stars). \textbf{(b)} Norm of the gradient without any credit through memory, relative to the full gradient, against the cost of removing all credit through memory. Gradient norms are medians over 64 probe batches from four separately trained full policies at update 1000, and costs are means over the 16 seeds of Section~\ref{sec:when_matters}. For the vessel, norms are medians over 46 windows at the final checkpoint, and costs are means over its 16 streams.}
\label{fig:gradients}
\end{figure}

To test whether a static gradient diagnostic could have predicted these costs without training ablations, we measured, at checkpoints of the full policy, the norm of each route's contribution relative to the norm of the full gradient (Figure~\ref{fig:gradients}). In the quadrotor and the vessel model, the size of a route's gradient predicted neither the truncation cost of removing it nor which of the two routes cost more.

Because the route contributions are vectors that sum to the full gradient and can partially cancel, a single route can be larger in norm than the full gradient itself. With noise of $0.05\text{ m/s}$, for example, the representation route was $1.74$ times the norm of the full gradient, yet removing it cost nothing ($-0.9\%$). Conversely, with velocity hidden the closed-loop route was more than twice the size of the representation route ($1.52$ vs.\ $0.68$), yet removing representation credit cost more. We found the same disconnect when all credit through memory was removed (Figure~\ref{fig:gradients}b) and in the vessel model, whose representation route carried $0.38$ of the gradient norm and cost $0.7\%$ to remove. We had also expected full physical credit to enlarge the vessel's closed-loop route more than its representation route, but this was not borne out (Appendix~\ref{app:gradients}).

Gradient norms capture the learning signal at a single moment of training, while the cost of a cut accumulates over the entire optimization trajectory. To decide which route to keep, a short comparison of training runs from the same initialization is therefore more informative than any static gradient diagnostic we tried.

\section{Discussion}
\label{sec:discussion}

\textbf{What we found.} A cut that leaves every forward computation unchanged, as the usual memory cut does, raised the final error of a quadrotor policy by a third when the policy could not observe its velocity, and cost nothing when the policy could. Within that cut, the two routes responded alike to observation noise and differently to physical credit: with velocity hidden, a mild decay of state-edge credit eliminated the truncation cost of closed-loop credit and left representation credit intact. Moreover, in a fixed-wing aircraft representation credit could stand in for closed-loop credit.

\textbf{When one route can stand in for the other.} We conjecture that one route can replace another to the extent that the two shape the same part of what the policy stores. Two observations from the quadrotor are consistent with this interpretation. First, the routes acted at different temporal lags there (Section~\ref{sec:span_dynamics}): closed-loop credit worked mainly through the one or two most recent stored steps (the positions that the latest commands have just moved), whereas representation credit drew on older steps, as it would if it shaped how a longer history of positions is combined against noise. Second, at low noise only closed-loop credit had measurable value (Section~\ref{sec:noise_both}): when velocity is nearly known, little needs to be combined over a longer history, so only the route that concerns the latest steps stays useful. On this reading, representation credit in the fixed-wing aircraft already reaches what closed-loop credit shapes there, suggesting a testable prediction: in that system, the span measurement of Section~\ref{sec:span_dynamics} should find closed-loop credit acting at lags where representation credit acts too.

\textbf{Implications for practice.} Table~\ref{tab:3} turns these observations into practical suggestions alongside the settings they rest on. The common thread, as we read it, is that the cost of cutting credit through memory depends on what the policy cannot observe and on whether the other route can take over. Neither can be read from gradient norms, so a short training comparison remains the most reliable guide.

\textbf{Connections to estimation and control.} These results echo a classical idea in stochastic control. In linear-quadratic-Gaussian control, the separation principle lets the state estimator be designed without regard to the control cost and lets the controller act on its estimate \citep{wonham1968separation}. End-to-end training of a memory policy imposes no such split, and our two routes map onto its two halves: representation credit parallels training the estimator for the task, and closed-loop credit parallels letting the controller account for how its actions shape the measurements the estimator will receive. Indeed, in the quadrotor with velocity hidden the second lost its value under a mild decay while the first kept it, which is consistent with a memory that works mainly as an estimator trained for the task. Costates offer a complementary view: closed-loop credit is the part of a physical state's costate (the sensitivity of the loss to that state) that arrives through memory. \citet{leeftink2026neural} find that costate structure in recurrent policies depends on how they are supervised, and our two routes split the supervision that memory receives by its source.

\begin{table}[t]
\caption{Suggestions for practice and the observations they rest on.}
\label{tab:3}
\begin{center}
\small
\begin{tabular}{P{0.19\linewidth}P{0.34\linewidth}P{0.22\linewidth}P{0.14\linewidth}}
\toprule
Situation & What we observed & Suggested practice & Basis \\
\midrule
State observed accurately & Cutting memory credit changed error by $<0.3\%$ (quadrotor), $<1.5\%$ (vessel) & Truncation is safe & Quadrotor, vessel \\
Velocity noisy or hidden, full physical credit & Cutting memory credit cost $13\text{--}34\%$ & Keep credit through memory & Quadrotor, 16 seeds \\
Velocity hidden, mild state-edge decay & Closed-loop truncation cost vanished; representation credit kept its value & If cutting, keep representation credit & Quadrotor, 16 new seeds \\
Closed-loop credit replaceable & Removing it alone cost almost nothing under every measure; removing both cost $5.8\%$ (median) & Cutting closed-loop credit may be harmless; check by training & Fixed-wing, 16 seeds \\
Unstable plant with full physical credit & Runs relapsed, and costs depended on when the error was measured & Stabilize training before comparing cuts & Cart-pole, 16 seeds \\
Detaching all observations & Error rose $51\text{--}107\%$, even with exact velocity & Avoid with memory policies at $\gamma=1$ & Quadrotor, 16 seeds \\
Choosing what to keep & Gradient norms did not predict truncation costs & Compare short training runs & Quadrotor and vessel \\
\bottomrule
\end{tabular}
\end{center}
\end{table}

\textbf{Partial cuts and numerical events.} The numerical events in the vessel runs (Section~\ref{sec:when_matters}) were unexpected, since truncation usually serves to tame exploding gradients \citep{metz2021gradients}. Here, removing only the part of the physical loop that runs through memory made training less stable. One hypothesis is that such a partial cut leaves a backward recursion whose terms no longer offset each other. Consistent with this, stopping gradients at every observation, which removes the loop altogether, produced no events. \citet{schnell2024stabilizing} point out that a modified backward pass no longer defines a gradient field, and these events may be one practical consequence. In the cart-pole, too, runs without closed-loop credit relapsed most often (Appendix~\ref{app:other_systems}), which we note only as consistent with the pattern.

\textbf{Limitations.} Our tasks are 32-step tracking problems with short memory needs, and the policies are small Transformers. Tasks that need long memory, or recurrent policies, may weigh the routes differently, and the arms apply to them without change. We observed the separation itself in one system: the closed-loop route could be replaced in the fixed-wing aircraft, and training in the cart-pole was too unstable to compare the routes, so a system in which closed-loop credit is indispensable but whose dynamics differ from the quadrotor's would show how general the separation is. Costs were also measured at a fixed budget of 1000 updates, and at 2000 updates the order of the two routes changed, so how they trade off over longer training remains open.

\section{Conclusion}
\label{sec:conclusion}

When does backpropagating through policy memory matter? In quadrotor tracking, it mattered when the policy could not observe its velocity reliably and cost nothing otherwise. Its two routes then behaved differently: a mild decay of credit through physical states eliminated the truncation cost of closed-loop credit and left representation credit intact, a separation that held on new seeds. In a fixed-wing aircraft closed-loop credit could be replaced by representation credit, which places the separation in systems where it cannot. Since gradient norms predicted none of this, the cost of cutting credit through memory is best measured by training. Which part of the memory each route shapes is the natural next question.

\subsubsection*{Data and code availability}

The code is in the supplementary material and will be released. The vessel data are commercial AIS records that cannot be shared, so the supplement provides summary statistics instead.

\bibliography{references}
\bibliographystyle{tmlr}

\newpage
\appendix
\renewcommand{\thetable}{\thesection\arabic{table}}
\makeatletter
\@addtoreset{table}{section}
\makeatother

\section*{Appendices}

The table below lists what each appendix contains and which part of the main text it supports.

\begin{center}
\small
\begin{tabular}{llP{0.42\linewidth}l}
\toprule
 & Title & Contents & Supports \\
\midrule
\ref{app:systems} & Systems and training & Dynamics, tasks, policies, optimizers, evaluation, and statistical methods for all four systems & Section~\ref{sec:setup} \\
\ref{app:predictions} & Predictions and their outcomes & The predictions we stated before the runs that test them, how each came out, why several were not established, and the choices made along the way & Sections~\ref{sec:setup}--\ref{sec:results} \\
\ref{app:quadrotor} & Quadrotor results & All truncation costs by condition, decay level, and seed set, with intervals, and the main comparisons under other measures of the final error & Sections~\ref{sec:when_matters}--\ref{sec:separation} \\
\ref{app:further} & Further differences & Span, training length, and switching experiments & Section~\ref{sec:span_dynamics} \\
\ref{app:other_systems} & Cart-pole and fixed-wing & Pilot runs, results under several measures of the final error, and linearizations & Sections~\ref{sec:other_systems}, \ref{sec:discussion} \\
\ref{app:vessel} & Vessel model & Learning results, clipping, and numerical events & Sections~\ref{sec:when_matters}, \ref{sec:discussion} \\
\ref{app:gradients} & Gradient measurements & Definitions, protocol, and all measured values & Section~\ref{sec:grad_results} \\
\ref{app:implementation} & Implementation and checks & How the arms are built and how we checked them & Sections~\ref{sec:routes}, \ref{sec:setup} \\
\ref{app:compute} & Compute & Hardware and run times & -- \\
\bottomrule
\end{tabular}
\end{center}

\section{Systems and training}
\label{app:systems}

This appendix gives the details behind Section~\ref{sec:setup}. Table~\ref{tab:A1} collects the settings that the four systems share or differ in.

\begin{table}[!ht]
\caption{Training settings. All policies are trained from the stated initialization with AdamW and no learning-rate schedule.}
\label{tab:A1}
\begin{center}
\small
\begin{tabular}{lllll}
\toprule
 & Quadrotor & Cart-pole & Fixed-wing & Vessel \\
\midrule
State, action dimensions & 12, 4 & 4, 1 & 12, 4 & 4, 2 \\
Control step, horizon & $0.05\text{ s}$, 32 & $0.05\text{ s}$, 32 & $0.05\text{ s}$, 32 & $5\text{ min}$, 288 \\
Policy & \multicolumn{3}{c}{2 layers, width 64, 4 heads} & $2\times 2$ layers, 256, 8 heads \\
Parameters & 94,324 & 93,129 & 93,940 & 3,677,698 \\
Initialization & random & random & random & pretrained \\
Learning rate & $10^{-3}$ & $10^{-3}$ & $5 \times 10^{-4}$ & $3 \times 10^{-5}$ \\
Adam $\beta$, weight decay & $(0.9, 0.999)$, $0.01$ & $(0.9, 0.999)$, $0.01$ & $(0.9, 0.999)$, $0.01$ & $(0.9, 0.99)$, $0.002$ \\
Batch size, updates & 8, 1000 & 8, 1000 & 8, 1000 & 4, 2400 \\
Gradient-norm clipping & 4311.6 & 4311.6 & 4311.6 & $2.41 \times 10^5$ \\
Evaluation panel & 256 traj. & 256 traj. & 256 traj. & 184 windows \\
Seeds or streams & 16 (+16) & 16 & 16 & 16 \\
\bottomrule
\end{tabular}
\end{center}
\end{table}

\textbf{Quadrotor.} The simulator builds on the quadrotor code of \citet{wiedemann2023apg} and keeps its force model, rate controller, and physical parameters, with rotor dynamics switched off. We changed the integrator in two places. Positions advance by $\Delta t\,v + \frac{1}{2}\Delta t^2 a$, and attitude advances by solving for the Euler-angle rates that correspond to the body rates; velocities and body rates use explicit Euler steps. The policy's four sigmoid outputs set a collective thrust acceleration of $15 a_0 - 7.5 + 9.81\text{ m/s}^2$ and three body-rate commands $a_i - 0.5$ that go to the original rate controller.

On each axis the reference velocity is $A \sin(\omega t + \phi)$ with $A \sim \mathcal{U}[0.1, 0.3]\text{ m/s}$, $\omega \sim \mathcal{U}[0.3, 0.8]\text{ rad/s}$, and $\phi \sim \mathcal{U}[0, 2\pi]$, and the reference position is its integral. Initial states are drawn uniformly per component: position within $\pm 0.1\text{ m}$, attitude within $\pm 0.04\text{ rad}$, velocity relative to the reference within $\pm 0.25\text{ m/s}$, and body rates within $\pm 0.05\text{ rad/s}$. At each step the policy reads a 15-dimensional token (attitude as a rotation matrix, angular velocity, and body-frame velocity) and a 10-step preview of the reference, encoded by a one-dimensional convolution, that contains the reference position relative to the vehicle, the reference velocity, and the difference between reference and current velocity. Because noise is added to the world-frame velocity, it reaches every velocity input. With velocity hidden, the body-frame velocity and the velocity difference are set to zero while the reference velocity, attitude, and angular velocity remain.

The policy is a causal Transformer with RMSNorm, rotary position embeddings, and SwiGLU feed-forward layers of width 128. Its cache holds the whole episode and starts empty. The training loss sums over the batch, time, and coordinates, with weights 10 on position error, 1 on velocity error, $0.1$ on angular velocity, 5 on the thrust command's deviation from $0.5$, and $0.1$ on the deviations of the rate commands. The clipping threshold is the median gradient norm of the full policy over 16 batches after 100 warm-up updates in a pilot run, and it is the same for all arms. Each seed fixes the initialization and the sequences of training batches and noise. Thus all arms of a seed see identical data. The replication in Section~\ref{sec:separation} uses 16 seeds that appear nowhere else, and the evaluation panel is the same for every run and every evaluation.

\textbf{Cart-pole.} We wrote a frictionless cart-pole for this study with cart mass $1\text{ kg}$, pole mass $0.1\text{ kg}$, pole half-length $0.5\text{ m}$, and $g = 9.81\text{ m/s}^2$. The action is a horizontal force of $30\tanh(u)\text{ N}$, and each $0.05\text{ s}$ control step consists of five semi-implicit Euler steps of $0.01\text{ s}$. Initial states are uniform within $\pm 0.5\text{ m}$, $\pm 0.5\text{ m/s}$, $\pm 0.3\text{ rad}$, and $\pm 0.5\text{ rad/s}$. The cart follows a reference made of two sinusoids with amplitudes $\mathcal{U}[0, 0.5]\text{ m}$, frequencies $\mathcal{U}[0.2, 0.8]\text{ Hz}$, and random phases. With exact velocity the policy observes $x - x_{\mathrm{ref}}$, $\sin\theta$, $\cos\theta$, $\dot{x}$, and $\dot{\theta}$ together with a preview of the reference. With velocity hidden, $\dot{x}$, $\dot{\theta}$, and the velocity difference in the preview are set to zero. The loss is the time average of $(x - x_{\mathrm{ref}})^2 + \theta^2 + 0.125\tanh(u)^2$, summed over the batch, and the final error drops the action term.

\textbf{Fixed-wing aircraft.} The aircraft follows the six-degree-of-freedom model and parameters in the code of \citet{wiedemann2023apg}, including its saturation of the angle of attack and the sideslip angle at $\pm 10^\circ$, and it uses the same integrator changes as the quadrotor. The four sigmoid outputs set a thrust of up to $7\text{ N}$, the elevator and rudder within $\pm 20^\circ$, and the ailerons within $\pm 2.5^\circ$. The aircraft starts at $11.5\text{ m/s}$, and its target is a point on a trajectory generated from random actions by the original sampler. The reference moves toward the target at a constant $12\text{ m/s}$. The policy observes the nine normalized non-position states and a preview of relative position, reference velocity, and velocity difference; with velocity hidden, the body-frame velocity and the velocity difference are zero. The loss is the time-averaged squared position error plus a small penalty on the actions' deviation from $0.5$.

\textbf{Vessel.} The data are AIS tracks of ten commercial vessels, divided into 329 segments that meet the quality criteria; 263 are used for training and 66 for calibration. Two further vessels are excluded from all training and selection. A training window is drawn by choosing a training segment uniformly and then a start time uniformly within it. The model reads a 16-step prefix, whose cache is built without gradients, and then rolls out 288 steps of 5 minutes. Its state is latitude, longitude, and the north and east velocities, and its action sets the next velocity, from which the next position follows on the sphere. The architecture of \citet{gan2026sketch} consists of two stacked Transformers. We load a backbone pretrained with teacher forcing, bypass its original output layer, and add a linear head, initialized to zero, whose output is added to the velocity at the start of the window. The training loss and the final error are the time-averaged squared position error in $\text{km}^2$.

We chose the learning rate, $3 \times 10^{-5}$, from four values between $3 \times 10^{-6}$ and $10^{-4}$ on 80 development windows from 20 calibration segments; the largest diverged. We trained for 2400 updates, by which point the development error had largely stopped improving. Clipping uses the 95th percentile of gradient norms early in training, because the 90th percentile clipped too often ($16\%$ of updates). The final error averages the remaining 184 calibration windows (46 segments, 4 windows each) over evaluations at updates 2000, 2200, and 2400, since a single evaluation was noisy on the oscillating development curves. Note that these windows come from segments held out from training but belong to the same ten vessels.

\textbf{Statistics.} For each seed, the truncation cost of an arm is $c = E_{\mathrm{arm}}/E_{\mathrm{full}} - 1$, where the full policy shares the seed, observation condition, and $\gamma$. The common-scale cost is $(E_{\mathrm{arm}}(\gamma) - E_{\mathrm{full}}(\gamma))/E_{\mathrm{full}}(1)$, and the interaction is $c_{\text{no-mem}} - c_{\text{no-rep}} - c_{\text{no-loop}}$. Intervals are percentile bootstrap intervals of the mean over seeds, or of the median over windows for the vessel gradient measurements, with 100,000 resamples. For predictions tested together on the same runs, intervals are widened for the number of predictions: to $98.3\%$ for the three tested on the original quadrotor seeds (Sections~\ref{sec:noise_both} and~\ref{sec:separation}) and to $97.5\%$ for the two tested on the new seeds. Table~\ref{tab:2} and the cart-pole and fixed-wing tables of Appendix~\ref{app:other_systems} also report medians, because a few seeds had very large costs. A run relapses if, after reaching its lowest error, its error rises to at least twice that value by update 1000, both read from the evaluations every 100 updates. The final error is the error at update 1000 everywhere except in Tables~\ref{tab:C5} and~\ref{tab:E3}, which recompute the main comparisons with five other measures taken from the same evaluations: the mean and the median over updates 800--1000 and over updates 600--1000, and the lowest error after update 500. We added these measures after seeing how often cart-pole runs relapsed. The sign test for the vessel events counts streams in which the arms without closed-loop credit had more events than the arms with it, ignoring ties.

\section{Predictions and their outcomes}
\label{app:predictions}

Several comparisons in Section~\ref{sec:results} were designed to test directional predictions that we recorded, together with their summary statistics and interval widths, before running the corresponding experiments. Table~\ref{tab:B1} lists these predictions and how each came out.

\begin{table}[!ht]
\caption{Predictions stated before the runs that test them. Differences are in percentage points with the widened intervals of Appendix~\ref{app:systems}.}
\label{tab:B1}
\begin{center}
\small
\begin{tabular}{P{0.46\linewidth}lP{0.40\linewidth}}
\toprule
Prediction & Where & Outcome \\
\midrule
Going from exact to noisy velocity raises the cost of removing representation credit more than that of removing closed-loop credit (quadrotor) & \ref{sec:noise_both} & Not supported: $-3.0$ $[-9.8, 3.0]$ \\
With velocity hidden, removing representation credit costs more than removing closed-loop credit & \ref{sec:noise_both} & Supported in the original seeds: $11.2$ $[0.1, 23.3]$, 13 of 16; close in the new seeds ($20\%$ and $19\%$) \\
Removing state-edge credit ($\gamma = 0$, noise $0.20$) lowers the closed-loop cost more than the representation cost & \ref{sec:separation} & Not supported: $3.4$ $[-2.7, 10.3]$; training stalled \\
Mild decay ($\gamma = 0.8$), new seeds, velocity hidden: lowers the closed-loop cost more than the representation cost & \ref{sec:separation} & Supported: $39$ $[10, 81]$, 14 of 16 \\
Mild decay ($\gamma = 0.8$), new seeds, noise $0.20$: lowers the closed-loop cost more than the representation cost & \ref{sec:separation} & Not established: $3.8$ $[-1.1, 9.1]$ \\
Keeping two recent steps recovers more of the representation cost than of the closed-loop cost & \ref{sec:span_dynamics} & Reversed at noise $0.20$ (ratio difference $-0.55$ $[-0.82, -0.34]$); not resolved with velocity hidden \\
Cart-pole, velocity hidden, decay to $\gamma = 0.9$: lowers the closed-loop cost more than the representation cost & \ref{sec:other_systems} & Not supported: positive in 10 of 16 seeds, interval includes zero \\
Fixed-wing aircraft, velocity hidden, decay to $\gamma = 0.9$: lowers the closed-loop cost more than the representation cost & \ref{sec:other_systems} & Not supported: $-6.5$ $[-36.8, 21.3]$, positive in 7 of 16 seeds \\
Vessel gradients: full physical credit enlarges the closed-loop share more than the representation share & \ref{sec:grad_results} & Not established (median $0.025$, $[-0.0005, 0.045]$); reversed at three other checkpoints \\
Vessel learning: removing closed-loop credit costs more than removing representation credit, and removing both costs more than nothing & \ref{sec:when_matters} & Clipping too frequent ($10\text{--}14\%$) for a clean comparison (Appendix~\ref{app:systems}); neither held \\
\bottomrule
\end{tabular}
\end{center}
\end{table}

Several of these contrasts were not established although the truncation costs behind them were large and consistent across seeds. A difference of differences is much noisier at 16 seeds than the costs it combines. Moreover, both single-route cuts remove the mixed paths, which carried $0.4$ to $1.2$ times the full gradient norm with noisy or hidden velocity (Table~\ref{tab:G1}), so the two costs tend to move together.

A few choices made along the way affect how the results should be read. The first test of state-edge decay used $\gamma = 0$, the strongest decay among our candidate levels under which short trial runs still learned. Gradient measurements on separately trained policies hinted that the noise prediction might fail, because the closed-loop route grew more with noise than the representation route; we kept the prediction as stated. The mild decay of $\gamma = 0.8$, in contrast, was chosen after the original seeds had been analyzed, which is why we tested it on new seeds.

In the vessel model, we had planned to treat the fine-tuning comparison as a formal test only if gradient clipping stayed rare ($<10\%$ of updates); because clipping remained active on $10\text{--}14\%$ of updates, we report the vessel learning results descriptively. The numerical events of Appendix~\ref{app:vessel} were noticed during the runs, so the check on further streams came afterwards, as described there.

\section{Quadrotor results}
\label{app:quadrotor}

Tables~\ref{tab:C1}--\ref{tab:C4} give every quadrotor truncation cost behind Sections~\ref{sec:when_matters}--\ref{sec:separation}. Two patterns in these tables are worth pointing out. First, the interaction of the two routes was small at $\gamma = 1$, and its intervals included zero in every condition. Second, a memory of two steps recovered most of what memory is worth with velocity hidden: its cost was $31\%$, against $107\%$ for a memory of one step. This is consistent with a velocity estimate that needs little more than the difference of two positions.

Table~\ref{tab:C4} also answers a question a practitioner may have. Once credit through earlier states is removed, stopping gradients at the observations as well costs nothing more, so the large observation cost of Section~\ref{sec:when_matters} belongs to training with full physical credit.

\begin{table}[!ht]
\caption{Quadrotor truncation costs at $\gamma = 1$ ($\%$), means with $95\%$ intervals over 16 seeds. The last row gives the full policy's median final error.}
\label{tab:C1}
\begin{center}
\small
\begin{tabular}{lllll}
\toprule
Arm & exact & noise 0.05 & noise 0.20 & hidden \\
\midrule
no representation credit & -0.2 [-0.4, 0.1] & -0.9 [-2.1, 0.2] & 12.6 [10.1, 15.2] & 26.1 [19.4, 33.9] \\
no closed-loop credit & 0.2 [-0.0, 0.4] & 6.3 [4.1, 8.4] & 16.0 [12.7, 19.9] & 14.9 [9.0, 22.2] \\
no credit through memory & -0.2 [-0.4, 0.0] & 3.3 [1.8, 4.8] & 26.9 [24.1, 29.8] & 33.9 [24.9, 43.5] \\
interaction & -0.2 [-0.5, 0.0] & -2.1 [-4.4, 0.3] & -1.8 [-5.5, 1.8] & -7.1 [-17.6, 3.7] \\
no observation gradient & 75.0 [44.6, 121.4] & 51.5 [41.2, 62.4] & 74.1 [66.1, 82.3] & 106.5 [90.4, 123.1] \\
memory of 1 step & -0.3 [-0.5, 0.0] & 2.7 [1.6, 3.6] & 38.4 [35.3, 41.4] & 106.8 [99.3, 113.8] \\
memory of 2 steps & -0.3 [-0.5, -0.0] & 1.3 [0.2, 2.3] & 32.3 [29.7, 35.0] & 30.5 [16.0, 48.2] \\
full-policy error ($10^{-3}$\,m$^2$) & 4.73 & 5.14 & 6.13 & 5.24 \\
\bottomrule
\end{tabular}
\end{center}
\end{table}

\begin{table}[!ht]
\caption{Quadrotor truncation costs ($\%$) under state-edge decay, original 16 seeds. Same-$\gamma$ costs are relative to the full policy at the same $\gamma$, and common-scale costs measure the excess error in units of the full policy's error at $\gamma = 1$. The full policy's error was $16$ and $18$ times its $\gamma = 1$ value at $\gamma = 0$ (noise $0.20$ and hidden velocity), and $1.7$ times at $\gamma = 0.8$.}
\label{tab:C2}
\begin{center}
\small
\begin{tabular}{lllll}
\toprule
Velocity & $\gamma$ & Arm & Same-$\gamma$ cost & Common-scale cost \\
\midrule
noise 0.20 & 0 & no rep. & 0.0 [-0.2, 0.3] & 0.5 [-3.8, 4.5] \\
 &  & no loop & 0.1 [-0.0, 0.2] & 1.4 [-0.6, 3.1] \\
 &  & no memory & 0.1 [-0.2, 0.4] & 1.6 [-2.9, 6.4] \\
 & 0.8 & no rep. & 7.6 [4.7, 10.6] & 13.0 [8.1, 18.2] \\
 &  & no loop & 5.4 [3.0, 7.9] & 9.1 [5.1, 13.4] \\
 &  & no memory & 13.1 [10.1, 16.4] & 22.7 [17.5, 28.4] \\
hidden & 0 & no rep. & 0.4 [0.1, 0.7] & 8.1 [2.8, 13.5] \\
 &  & no loop & -0.0 [-0.1, 0.1] & -0.3 [-1.9, 1.2] \\
 &  & no memory & 0.4 [0.1, 0.6] & 6.8 [1.8, 11.5] \\
 & 0.8 & no rep. & 33.9 [28.0, 39.6] & 56.5 [46.8, 66.1] \\
 &  & no loop & -1.4 [-4.3, 1.1] & -2.7 [-7.8, 1.7] \\
 &  & no memory & 27.0 [22.6, 31.6] & 44.7 [37.9, 51.8] \\
\bottomrule
\end{tabular}
\end{center}
\end{table}

\begin{table}[!ht]
\caption{Quadrotor truncation costs ($\%$) in the 16 new seeds. $^\dagger 97.5\%$ interval, widened as described in Appendix~\ref{app:systems}.}
\label{tab:C3}
\begin{center}
\small
\begin{tabular}{lllll}
\toprule
Velocity & $\gamma$ & Arm & Same-$\gamma$ cost & Common-scale cost \\
\midrule
hidden & 1 & no rep. & 20.2 [13.4, 27.5] & -- \\
 &  & no loop & 18.5 [6.1, 39.7] & -- \\
 &  & no memory & 37.3 [29.3, 46.1] & -- \\
 & 0.8 & no rep. & 40.8 [24.0, 69.0] & 65.8 [39.4, 110.1] \\
 &  & no loop & -0.4 [-2.1, 1.5] & -0.6 [-3.5, 2.5] \\
 &  & no memory & 24.8 [18.6, 30.9] & 40.1 [30.4, 49.3] \\
 &  & difference of drops & 39.4 [10.5, 81.3]$^\dagger$ & median 14.6; 14/16 positive \\
noise 0.20 & 1 & no rep. & 18.3 [15.4, 21.1] & -- \\
 &  & no loop & 16.8 [15.0, 18.6] & -- \\
 &  & no memory & 33.2 [29.8, 36.9] & -- \\
 & 0.8 & no rep. & 8.6 [4.8, 12.8] & 15.0 [8.4, 22.2] \\
 &  & no loop & 3.3 [-1.6, 7.1] & 4.9 [-6.0, 12.6] \\
 &  & no memory & 8.9 [3.7, 13.8] & 15.1 [5.1, 23.9] \\
 &  & difference of drops & 3.8 [-1.1, 9.1]$^\dagger$ & median 2.9; 10/16 positive \\
\bottomrule
\end{tabular}
\end{center}
\end{table}

\begin{table}[!ht]
\caption{One-step physical credit combined with no observation gradient, the configuration closest to the controllers of \citet{nachkov2025controllers}. Costs ($\%$) are relative to the full policy at $\gamma = 0$; the last column gives how much worse the full policy is at $\gamma = 0$ than at $\gamma = 1$.}
\label{tab:C4}
\begin{center}
\small
\begin{tabular}{llll}
\toprule
Velocity & Cost vs.\ full policy at $\gamma=0$ & Common-scale cost & Full policy, $\gamma=0$ / $\gamma=1$ \\
\midrule
exact & 0.0 [-0.1, 0.1] & 0.3 [-1.7, 2.3] & 20.3 \\
noise 0.05 & -0.0 [-0.1, 0.1] & -0.1 [-1.8, 1.5] & 18.7 \\
noise 0.20 & 0.1 [-0.0, 0.2] & 1.7 [-0.4, 3.6] & 16.1 \\
hidden & -0.0 [-0.1, 0.1] & -0.4 [-2.3, 1.6] & 18.5 \\
\bottomrule
\end{tabular}
\end{center}
\end{table}

\textbf{Other measures of the final error.} Table~\ref{tab:C5} recomputes the main quadrotor comparisons with the five other measures of Appendix~\ref{app:systems}, and the comparisons it covers point the same way under every measure. The mean difference of drops in the new seeds is the one number that depends strongly on the measure, ranging from 17 to 109 points, because two seeds dominate it. In one, the error of the arm without closed-loop credit at $\gamma = 1$ spiked near update 900, and in the other, the error of the arm without representation credit at $\gamma = 0.8$ jumped at update 1000. The median difference stays between 15 and 22 points, and the difference is positive in 14 to 16 of the 16 seeds.

\begin{table}[!ht]
\caption{Quadrotor truncation costs ($\%$) under six measures of the final error (Appendix~\ref{app:systems}). Means over 16 seeds, with the number of seeds in which the error increased. For the difference of drops from $\gamma = 1$ to $\gamma = 0.8$, the mean and the median, with the number of seeds in which the difference was positive. The first measure is the one used in the main text.}
\label{tab:C5}
\begin{center}
\footnotesize
\setlength{\tabcolsep}{3.5pt}
\begin{tabular}{llrrrrrr}
\toprule
 &  & at & mean & median & mean & median & lowest \\
Velocity & Arm & 1000 & 800--1000 & 800--1000 & 600--1000 & 600--1000 & 500--1000 \\
\midrule
\multicolumn{8}{l}{\emph{Original 16 seeds, $\gamma = 1$}} \\
hidden & no rep. & 26.1 (16) & 25.3 (16) & 25.3 (16) & 25.0 (16) & 25.4 (16) & 20.7 (16) \\
 & no loop & 14.9 (14) & 17.4 (16) & 13.9 (16) & 15.3 (16) & 10.9 (16) & 8.0 (16) \\
 & no memory & 33.9 (16) & 38.8 (16) & 40.4 (16) & 39.1 (16) & 37.6 (16) & 28.2 (16) \\
noise 0.20 & no rep. & 12.6 (16) & 11.3 (16) & 11.0 (15) & 9.0 (16) & 10.2 (16) & 12.1 (16) \\
 & no loop & 16.0 (16) & 17.3 (16) & 17.9 (16) & 16.5 (16) & 18.0 (16) & 14.7 (16) \\
 & no memory & 26.9 (16) & 25.6 (16) & 24.6 (16) & 22.2 (16) & 24.1 (16) & 26.2 (16) \\
\multicolumn{8}{l}{\emph{New 16 seeds, $\gamma = 0.8$, and the difference of drops from $\gamma = 1$}} \\
hidden & no rep. & 40.8 (16) & 36.1 (16) & 31.4 (16) & 35.1 (16) & 31.6 (16) & 28.3 (16) \\
 & no loop & -0.4 (5) & -0.9 (5) & -1.3 (4) & -1.4 (4) & -1.7 (5) & -0.5 (7) \\
 & difference, mean & 39.4 & 108.9 & 29.7 & 73.3 & 20.3 & 16.7 \\
 & difference, median & 14.6 (14) & 18.5 (15) & 16.4 (16) & 22.4 (16) & 19.5 (16) & 14.9 (15) \\
noise 0.20 & no rep. & 8.6 (14) & 8.2 (15) & 8.3 (16) & 7.2 (16) & 8.4 (16) & 7.5 (15) \\
 & no loop & 3.3 (12) & 4.9 (13) & 5.8 (14) & 3.9 (15) & 4.9 (15) & 5.3 (13) \\
 & difference, mean & 3.8 & 5.3 & 3.6 & 8.1 & 7.7 & 3.1 \\
 & difference, median & 2.9 (10) & 7.3 (12) & 4.9 (11) & 8.8 (15) & 7.2 (12) & 5.3 (11) \\
\bottomrule
\end{tabular}
\end{center}
\end{table}

\section{Further differences between the routes}
\label{app:further}

These experiments support Section~\ref{sec:span_dynamics}. All use noise of $0.20\text{ m/s}$ unless stated, and their truncation costs are relative to the full policy of Table~\ref{tab:C1}.

\textbf{Span.} In these runs, credit through memory was kept for the $s$ most recent stored steps and removed for older ones (Table~\ref{tab:D1}). With $s = 0$ this is the full cut, and with $s = 32$ it would be the full policy. At noise $0.20\text{ m/s}$, removing all credit through memory cost roughly the sum of the two single-route costs at every span.

\begin{table}[!ht]
\caption{Truncation cost ($\%$) when credit through memory is kept for the $s$ most recent stored steps, for noise $0.20\text{ m/s}$ (top) and hidden velocity (bottom). Means with $95\%$ intervals over 16 seeds.}
\label{tab:D1}
\begin{center}
\small
\begin{tabular}{llll}
\toprule
Steps kept & No rep. & No loop & No memory \\
\midrule
0 (full cut) & 12.6 [10.1, 15.2] & 16.0 [12.6, 20.0] & 26.9 [24.1, 29.9] \\
1 & 11.9 [9.8, 14.0] & 8.7 [6.5, 11.2] & 21.8 [18.3, 25.7] \\
2 & 11.5 [9.3, 13.8] & 5.9 [4.0, 7.8] & 17.3 [14.5, 20.4] \\
4 & 4.3 [2.5, 5.8] & 2.1 [0.8, 3.5] & 8.1 [6.5, 9.7] \\
8 & 0.5 [-0.8, 1.8] & 0.5 [-0.4, 1.4] & 1.3 [0.4, 2.3] \\
16 & -0.3 [-0.8, 0.1] & -0.1 [-0.4, 0.2] & -0.3 [-0.8, 0.1] \\
\midrule
Steps kept & No rep. & No loop & No memory \\
\midrule
0 (full cut) & 26.1 [19.4, 34.0] & 14.9 [9.0, 22.1] & 33.9 [24.9, 43.6] \\
1 & 21.5 [16.2, 26.7] & 12.9 [7.7, 18.5] & 31.7 [23.1, 40.4] \\
2 & 10.7 [7.9, 13.6] & 3.1 [1.0, 5.7] & 17.8 [10.8, 27.3] \\
4 & 1.8 [0.1, 3.8] & 0.6 [-0.5, 1.8] & 2.5 [0.8, 4.3] \\
8 & -0.3 [-1.3, 0.7] & 0.1 [-0.6, 0.8] & -0.8 [-1.7, 0.1] \\
16 & 0.3 [-0.4, 1.1] & 0.4 [-0.4, 1.1] & 0.9 [-0.2, 2.2] \\
\bottomrule
\end{tabular}
\end{center}
\end{table}

\textbf{Training length.} We trained 8 of the 16 seeds for 2000 updates and recorded the error at updates 1000 and 2000 of the same runs (Table~\ref{tab:D2}). The two single-route costs swapped order. The cost of removing all credit through memory fell as well, though less than the closed-loop cost.

\begin{table}[!ht]
\caption{Truncation cost ($\%$) after 1000 and 2000 updates of the same runs, 8 seeds.}
\label{tab:D2}
\begin{center}
\small
\begin{tabular}{lllll}
\toprule
Arm & 1000 updates & 2000 updates & Change & Seeds with increase \\
\midrule
no rep. & 13.9 [10.0, 18.0] & 22.0 [15.6, 28.8] & 8.1 [-1.4, 17.0] & 5/8 \\
no loop & 14.8 [10.4, 19.9] & 5.8 [3.5, 8.0] & -8.9 [-13.4, -4.4] & 2/8 \\
no memory & 27.8 [25.6, 30.3] & 21.3 [16.4, 25.8] & -6.5 [-11.9, -0.9] & 3/8 \\
\bottomrule
\end{tabular}
\end{center}
\end{table}

\textbf{Switching.} In these runs the full policy trained until update $\tau$ and then continued with a cut, either keeping the optimizer's state or resetting it (Table~\ref{tab:D3}). However, resetting the optimizer alone, without any cut, lowered the error by $7.2\%$ and $5.5\%$ when done at updates 250 and 500. These reset arms therefore mix two effects, so the main text uses the arms that kept the optimizer state. Table~\ref{tab:D4} reverses the switch. When the representation cut was lifted at update 500, the policy caught up with the full policy, while lifting the closed-loop cut left a $6.3\%$ cost as long as the optimizer state was kept. Resetting the optimizer removed most of that difference, which suggests that part of the closed-loop cut's lasting cost was carried by the optimizer's state rather than by the weights.

\begin{table}[!ht]
\caption{Truncation cost ($\%$) when the cut starts at update $\tau$, with the optimizer state carried over or reset. The last column resets the optimizer without any cut.}
\label{tab:D3}
\begin{center}
\small
\begin{tabular}{llllll}
\toprule
Switch at & No rep., carried & No rep., reset & No loop, carried & No loop, reset & Reset only \\
\midrule
250 & 13.8 [11.7, 16.7] & 7.6 [4.0, 11.9] & 14.7 [11.6, 18.2] & 12.6 [9.4, 17.0] & -7.2 [-8.7, -5.8] \\
500 & 9.8 [8.6, 11.2] & 8.1 [6.2, 10.3] & 9.6 [7.4, 11.8] & 8.2 [5.7, 11.0] & -5.5 [-6.9, -4.3] \\
750 & 4.8 [2.9, 6.8] & 5.4 [3.7, 7.0] & 5.5 [3.5, 7.7] & 5.7 [4.9, 6.6] & 0.3 [-1.1, 1.6] \\
\bottomrule
\end{tabular}
\end{center}
\end{table}

\begin{table}[!ht]
\caption{Truncation cost ($\%$) when a cut that ran from the start is lifted at update 500.}
\label{tab:D4}
\begin{center}
\small
\begin{tabular}{lll}
\toprule
Switch at update 500 & Optimizer state carried & Optimizer state reset \\
\midrule
no rep. $\to$ full & -1.1 [-2.9, 0.6] & -4.9 [-6.7, -3.2] \\
no loop $\to$ full & 6.3 [4.6, 7.9] & -4.0 [-5.5, -2.7] \\
\bottomrule
\end{tabular}
\end{center}
\end{table}

\section{Cart-pole and fixed-wing aircraft}
\label{app:other_systems}

The two further systems of Section~\ref{sec:other_systems} use the policy architecture and arms of the quadrotor (Appendix~\ref{app:systems}). Here we give their pilots, their results, and the linearizations behind Section~\ref{sec:other_systems}.

\textbf{Pilot runs.} In each system we first ran short pilots on separate seeds to choose the learning rate and the decay level. In the cart-pole we compared three learning rates and three decay levels, and took $10^{-3}$, the best rate that trained stably, with $\gamma = 0.9$, the strongest decay that kept the full policy's error within $2.5$ times its value at $\gamma = 1$ ($1.5$ times, against $3.5$ times at $\gamma = 0.8$). The pilots also showed that removing all credit through memory had a clear cost ($8\%$), which a comparison of the two routes needs. The same procedure gave a learning rate of $5 \times 10^{-4}$ and $\gamma = 0.9$ for the fixed-wing aircraft. There, however, the pilots on eight seeds suggested that closed-loop credit had little truncation cost to remove ($1.1\%$ on average), and the decay comparison is informative only if it has some. Although we had initially considered skipping the full fixed-wing decay comparison unless its pilot closed-loop cost reached about $5\%$, we ultimately ran all 16 seeds so that both additional systems are reported on an equal footing.

\textbf{Cart-pole results.} Table~\ref{tab:E1} gives the cart-pole truncation costs over all 16 seeds, together with the number of runs in each arm that relapsed. The heavy tails come from these relapses. A run whose error happened to be rising at update 1000 shows a cost of hundreds or thousands of percent, although earlier in training its error had been close to that of the other runs. Table~\ref{tab:E3} shows that the medians, and the order of the routes, change with the measure of final error. At $\gamma = 0.9$ fewer runs relapsed, 12 of 64 against 23 of 64 at $\gamma = 1$. Giving the policy exact velocity made matters worse: the full policy ended worse in 14 of 16 seeds (median error $5.36$ against $0.17$ with velocity hidden), gradient clipping was active on $32\%$ of updates, and raw gradient norms reached about $2 \times 10^{10}$. The linearization explains why. Near the upright position the pole angle obeys $\ddot{\theta} = \lambda^2 \theta$ with
\begin{equation}
\label{eq:cartpole_lambda}
\lambda = \sqrt{\frac{g}{\ell\bigl(4/3 - m/(M + m)\bigr)}} = 3.97\text{ s}^{-1},
\end{equation}
where $\ell$ is the half-length and $m$ and $M$ are the pole and cart masses. One control step of the integrator multiplies this mode by $1.22$, so it grows by a factor of about $580$ over 32 steps. A decay of $\gamma = 0.9$ brings the factor per step to $1.10$, which may be why it trained stably in the pilots. Neither the cart-pole nor the fixed-wing aircraft produced numerical events of the kind seen in the vessel model (Appendix~\ref{app:vessel}).

\begin{table}[!ht]
\caption{Cart-pole truncation costs ($\%$), velocity hidden, all 16 seeds. Means with $95\%$ intervals, and medians with the number of seeds in which the error increased. The last column counts the runs in each arm that relapsed (Appendix~\ref{app:systems}).}
\label{tab:E1}
\begin{center}
\small
\begin{tabular}{lllll}
\toprule
$\gamma$ & Arm & Mean [95\% interval] & Median (seeds with increase) & Runs that relapsed \\
\midrule
1 & full policy & -- & -- & 3/16 \\
 & no rep. & 161.5 [-13.6, 494.3] & 2.9 (9/16) & 3/16 \\
 & no loop & 860.4 [8.9, 2515.6] & 9.1 (9/16) & 9/16 \\
 & no memory & 45.1 [-0.7, 100.7] & 9.3 (12/16) & 8/16 \\
0.9 & full policy & -- & -- & 2/16 \\
 & no rep. & 37.0 [6.2, 83.5] & 9.8 (13/16) & 4/16 \\
 & no loop & 62.7 [-14.0, 200.1] & 0.8 (10/16) & 3/16 \\
 & no memory & 171.2 [-8.0, 505.3] & 3.7 (10/16) & 3/16 \\
\bottomrule
\end{tabular}
\end{center}
\end{table}

\textbf{Fixed-wing results.} Table~\ref{tab:E2} gives the costs of the 16-seed comparison, and Table~\ref{tab:E3} gives the medians under the other measures of final error. At $\gamma = 1$, removing closed-loop credit alone kept the median error within $0.3\%$ of the full policy's under every measure, although two seeds had costs above $20\%$ at update 1000. Removing both routes raised the median error by $1.3\%$ to $5.8\%$, depending on the measure, and removing representation credit alone by $0.9\%$ to $4.9\%$. The pilots on eight other seeds gave similar medians. Unlike in the quadrotor and the cart-pole, the decay of $\gamma = 0.9$ did not slow training at all: the full policy ended at $0.91$ times its error at $\gamma = 1$, and no update was clipped. At update 1000 it did, however, raise the cost of every cut, closed-loop credit included, so the drop that the comparison looks for did not occur, and the difference between the drops was negative on average and positive in only 7 of the 16 seeds. With exact velocity, the full policy's error was $40\text{--}49\%$ of its error with velocity hidden (four pilot seeds), so velocity information mattered here as well.

Indeed, the aircraft itself is benign near level flight. Linearized at a trimmed flight of $11.5\text{ m/s}$ (angle of attack $0.84^\circ$, thrust $1.84\text{ N}$), all modes of the dynamics apart from the neutral position and heading are stable. The slowest is an airspeed--pitch oscillation with a period of $5.2\text{ s}$ and a damping ratio of $0.085$, about three times longer than an episode.

\begin{table}[!ht]
\caption{Fixed-wing truncation costs ($\%$), velocity hidden, 16 seeds. Means with $95\%$ intervals, and medians with the number of seeds in which the error increased.}
\label{tab:E2}
\begin{center}
\small
\begin{tabular}{llll}
\toprule
$\gamma$ & Arm & Mean [95\% interval] & Median (seeds with increase) \\
\midrule
1 & no rep. & 7.2 [1.5, 13.3] & 4.9 (11/16) \\
 & no loop & 3.7 [0.3, 8.0] & 0.3 (10/16) \\
 & no memory & 8.7 [2.8, 15.3] & 5.8 (11/16) \\
 & interaction & -2.2 [-5.8, 0.6] & 0.0 (8/16) \\
0.9 & no rep. & 18.1 [0.8, 39.0] & 12.3 (10/16) \\
 & no loop & 21.2 [-1.0, 52.8] & 6.1 (10/16) \\
 & no memory & 18.3 [1.1, 37.0] & 8.7 (11/16) \\
 & interaction & -21.0 [-59.2, 7.6] & -1.1 (8/16) \\
 & difference of drops & -6.5 [-36.8, 21.3] & -9.2 (7/16) \\
\bottomrule
\end{tabular}
\end{center}
\end{table}

\begin{table}[!ht]
\caption{Fixed-wing and cart-pole truncation costs ($\%$) at $\gamma = 1$ under six measures of the final error (Appendix~\ref{app:systems}), velocity hidden. Medians over 16 seeds, with the number of seeds in which the error increased. The last row of each system gives the difference of drops from $\gamma = 1$ to $\gamma = 0.9$, with the number of seeds in which it was positive. The first measure is the one used in the main text.}
\label{tab:E3}
\begin{center}
\footnotesize
\setlength{\tabcolsep}{3.5pt}
\begin{tabular}{llrrrrrr}
\toprule
 &  & at & mean & median & mean & median & lowest \\
System & Arm & 1000 & 800--1000 & 800--1000 & 600--1000 & 600--1000 & 500--1000 \\
\midrule
fixed-wing & no rep. & 4.9 (11) & 1.4 (9) & 0.9 (10) & 1.1 (11) & 1.6 (12) & 2.0 (10) \\
 & no loop & 0.3 (10) & 0.1 (11) & -0.2 (6) & 0.2 (10) & 0.1 (9) & 0.2 (11) \\
 & no memory & 5.8 (11) & 2.5 (10) & 2.2 (12) & 1.3 (12) & 3.0 (13) & 4.1 (12) \\
 & difference of drops & -9.2 (7) & 2.2 (8) & 2.2 (9) & -0.2 (8) & 5.2 (10) & -7.1 (6) \\
cart-pole & no rep. & 2.9 (9) & 9.1 (10) & 8.4 (10) & -0.6 (8) & 0.4 (8) & 3.8 (11) \\
 & no loop & 9.1 (9) & 14.9 (10) & 7.0 (10) & 48.1 (14) & 6.9 (11) & 2.3 (10) \\
 & no memory & 9.3 (12) & 21.3 (10) & 3.9 (12) & 26.0 (10) & -1.7 (7) & 3.3 (12) \\
 & difference of drops & 22.8 (10) & 48.9 (10) & 13.1 (10) & 86.7 (12) & 9.5 (12) & 5.0 (12) \\
\bottomrule
\end{tabular}
\end{center}
\end{table}

\section{Vessel model}
\label{app:vessel}

Table~\ref{tab:F1} gives the vessel's learning truncation costs, clipping rates, and numerical events by arm, which Sections~\ref{sec:when_matters} and~\ref{sec:discussion} summarize.

\begin{table}[!ht]
\caption{Vessel model, 16 sampling streams. Truncation costs ($\%$) average the final errors at updates 2000, 2200, and 2400. Clipping is the mean share of clipped updates per run, with its range. Events are numerical events in the six streams where we first noticed the pattern and in the ten further streams.}
\label{tab:F1}
\begin{center}
\footnotesize
\setlength{\tabcolsep}{4pt}
\begin{tabular}{lllllll}
\toprule
Arm & Cost [95\% interval] & Median & Increase & Clipped ($\%$) & Events, first six & Events, further ten \\
\midrule
full policy & -- & -- & -- & 11.4 (10.4--12.9) & 6 & 0 \\
no rep. & 0.7 [0.0, 1.2] & 1.0 & 12/16 & 10.2 (9.1--11.2) & 2 & 5 \\
no loop & -0.3 [-1.2, 0.7] & -0.4 & 6/16 & 12.5 (11.1--13.5) & 22 & 78 \\
no memory & 0.4 [-0.6, 1.5] & -0.5 & 6/16 & 11.0 (9.9--12.7) & 79 & 64 \\
no obs.\ gradient & -0.4 [-1.3, 0.4] & -0.3 & 7/16 & 13.9 (12.5--15.2) & 0 & 0 \\
\bottomrule
\end{tabular}
\end{center}
\end{table}

\textbf{Numerical events.} We handle two kinds of numerical event, and all reported runs use these rules. If every gradient entry is finite but the total norm overflows single precision, we recompute the norm in double precision and clip as usual. If a gradient entry is not finite, we skip the update. We added these rules after early runs failed on such values, and 18 of the 21 failed runs were in arms without closed-loop credit. All runs in Table~\ref{tab:F1} were trained from the beginning with the rules in place.

Of the 256 events in these runs, 196 were norm overflows and 60 were skipped updates, and the arms without closed-loop credit account for 185 and 58 of them. We first noticed the pattern in the six streams of Table~\ref{tab:F1}, where these arms produced 101 events against 8. In the ten further streams, these arms had more events than the arms with closed-loop credit in all nine streams that had any (one-sided sign test, $p = 0.002$). However, two of those streams had been looked at earlier, while we examined the failed runs; without them the count was 7 of 7 ($p = 0.008$).

\section{Gradient measurements}
\label{app:gradients}

The gradient measurements of Section~\ref{sec:grad_results} work as follows. On a given batch we compute the gradients of all four arms at the same parameters and recover the path classes of Equation~\eqref{eq:four_arms} as $P_{\mathrm{rep}} = g_{\text{no-loop}} - g_{\text{no-mem}}$, $P_{\mathrm{loop}} = g_{\text{no-rep}} - g_{\text{no-mem}}$, and $P_{\mathrm{mix}} = g_{\mathrm{full}} - g_{\text{no-rep}} - g_{\text{no-loop}} + g_{\text{no-mem}}$. These norms are reported relative to $\|g_{\mathrm{full}}\|$.

For each quadrotor condition, we trained the full policy on four separate seeds, saved checkpoints at updates 200, 500, and 1000, and evaluated 16 fresh batches at each checkpoint. Table~\ref{tab:G1} gives medians over the 64 batches at update 1000.

\begin{table}[!ht]
\caption{Quadrotor gradient norms relative to the full gradient, medians over 64 probe batches at update 1000.}
\label{tab:G1}
\begin{center}
\small
\begin{tabular}{llllll}
\toprule
$\gamma$ & Velocity & $\|P_{\mathrm{rep}}\|/\|g\|$ & $\|P_{\mathrm{loop}}\|/\|g\|$ & $\|P_{\mathrm{mix}}\|/\|g\|$ & $\|g_{\text{no-mem}}\|/\|g\|$ \\
\midrule
1 & exact & 0.21 & 0.11 & 0.02 & 0.94 \\
 & noise 0.05 & 1.74 & 0.81 & 1.11 & 1.27 \\
 & noise 0.20 & 0.66 & 0.66 & 0.45 & 1.09 \\
 & hidden & 0.68 & 1.52 & 1.19 & 0.93 \\
0 & exact & 0.21 & 0.04 & 0.01 & 1.02 \\
 & noise 0.05 & 0.21 & 0.05 & 0.01 & 1.04 \\
 & noise 0.20 & 0.16 & 0.04 & 0.01 & 1.04 \\
 & hidden & 0.51 & 0.01 & 0.00 & 1.12 \\
\bottomrule
\end{tabular}
\end{center}
\end{table}

The vessel measurements use the 46 calibration segments, one window each, at the checkpoints listed in Table~\ref{tab:G2}. Besides the residual parameterization used throughout the paper, the table includes the direct-output version of the vessel model that we worked with before adding the zero-initialized residual head of Appendix~\ref{app:systems}; it predicts velocity directly from the pretrained backbone, and we measured it both as pretrained and after two runs of further training with full physical credit. We include both parameterizations because the vessel prediction of Appendix~\ref{app:predictions} came out differently on the two versions, and because they differ in which route dominates. The prediction of Appendix~\ref{app:predictions} concerns the statistic
\begin{equation}
D = \log\!\left(S_{\mathrm{loop}}^{\mathrm{full}} / S_{\mathrm{loop}}^{\mathrm{one}}\right) - \log\!\left(S_{\mathrm{rep}}^{\mathrm{full}} / S_{\mathrm{rep}}^{\mathrm{one}}\right),
\end{equation}
where $S$ is a route's norm relative to the full gradient, measured with full physical credit or with credit through states cut after one step, the observation path kept in both. A positive $D$ thus means that full physical credit enlarges the closed-loop route by a larger factor than the representation route. In the residual model $D$ was small and positive at every checkpoint, with an interval that reaches zero at the main one, and in the direct-output model it was negative at all three. The two parameterizations also differ in which route dominates: the direct-output model's gradient through memory runs mostly along the closed-loop route, and the residual model's mostly along the representation route. Indeed, on the pretrained direct-output model, cutting all memory paths, including the gradients that the key and value weights receive from earlier positions, shrank the gradient to about a tenth of its norm under full physical credit and left it nearly unchanged when credit through states was cut after one step (32 windows). The vessel's mixed paths were at the level of rounding error ($\|P_{\mathrm{mix}}\|/\|g_{\mathrm{full}}\| \approx 6 \times 10^{-7}$), so we do not claim they are nonzero.

\begin{table}[!ht]
\caption{Vessel gradient norms with full physical credit (medians over 46 windows) and the statistic $D$ with a $95\%$ interval.}
\label{tab:G2}
\begin{center}
\small
\begin{tabular}{lllll}
\toprule
Checkpoint & $\|P_{\mathrm{rep}}\|/\|g\|$ & $\|P_{\mathrm{loop}}\|/\|g\|$ & $D$, median [95\% interval] & Windows \\
\midrule
residual model, 200 updates & 0.022 & 0.012 & 0.018 [0.014, 0.025] & 46 \\
residual model, 500 updates & 0.069 & 0.039 & 0.031 [0.017, 0.040] & 46 \\
residual model, 2400 updates (main) & 0.376 & 0.120 & 0.025 [-0.000, 0.045] & 46 \\
direct-output model, pretrained & 0.003 & 0.902 & -0.146 [-0.260, -0.046] & 46 \\
direct-output model, fine-tuned A & 0.006 & 0.812 & -0.372 [-0.406, -0.290] & 46 \\
direct-output model, fine-tuned B & 0.006 & 0.840 & -0.337 [-0.396, -0.252] & 46 \\
\bottomrule
\end{tabular}
\end{center}
\end{table}

\section{Implementation and checks}
\label{app:implementation}

The arms of Section~\ref{sec:routes} are built as follows. Each attention layer keeps two copies of its stored representations, a live copy $n_{\mathrm{live}}$ that the current step uses and a stored copy $n_{\mathrm{store}}$ that later steps read. The arms differ only in how the stored copy is computed (Section~\ref{sec:routes}), and later steps project it with the current key and value weights. This also makes the span experiments simple: restricting the cut to the $s$ most recent steps reads the live copy for those steps. Algorithm~1 gives one step of a rollout as pseudocode. The no-observation-gradient arm instead detaches $o_j$ in the computation that acts, and keeps $n_{\mathrm{store}} = n_{\mathrm{live}}$.

Before the main runs we checked the implementation in three ways. All arms produce the same rollouts as the full policy, and when credit is kept over the whole episode their gradients equal its gradient to numerical precision. The identities of Equation~\eqref{eq:four_arms} hold where they must; with observations detached from the state, for example, removing closed-loop credit changes nothing. Finally, on a small hand-built graph, the four path classes computed by automatic differentiation agree with those assembled from finite-difference Jacobians. We also confirmed that, in the vessel model, all three path classes vanish at initialization, where the zero-initialized head blocks every route.

The quadrotor and fixed-wing simulators build on the code of \citet{wiedemann2023apg} with the changes described in Appendix~\ref{app:systems}; the cart-pole and the arms are our own code. The supplementary material contains the code for all experiments.

\section{Compute}
\label{app:compute}

All experiments ran on a single server. Quadrotor, cart-pole, and fixed-wing runs each used one CPU core of an AMD EPYC 9654, and vessel runs used NVIDIA RTX 4090 GPUs, several runs to a GPU. Training and evaluation took about 860 hours of run time in total, about 560 of them for the vessel, 190 for the quadrotor, 30 for the fixed-wing aircraft, and 27 for the cart-pole. Because runs shared devices, these hours overstate the exclusive device time.

\end{document}